\documentclass[10pt,twocolumn,letterpaper]{article}

\usepackage{cvpr}              
\definecolor{cvprblue}{rgb}{0.21,0.49,0.74}
\usepackage[pagebackref,breaklinks,colorlinks,allcolors=cvprblue]{hyperref}
\usepackage{multirow}

\def\confName{CVPR}
\def\confYear{2026}

\title{Seeing Beyond Haze: Generative Nighttime Image Dehazing}

\author{Beibei Lin$^{1}$\hspace{0.1in}
Stephen Lin$^{2}$\hspace{0.1in}
Robby Tan$^{1}$
\\
$^1${National University of Singapore}, 
$^2${Microsoft Research Asia} 
\\
{\tt\small beibei.lin@u.nus.edu, stevelin@microsoft.com, robby.tan@nus.edu.sg}
}

\begin{document}

\twocolumn[{%
	\maketitle
	\renewcommand\twocolumn[1][]{#1}%
	\begin{center}
		\centering
		\includegraphics[width=\linewidth, height = 0.60\linewidth]{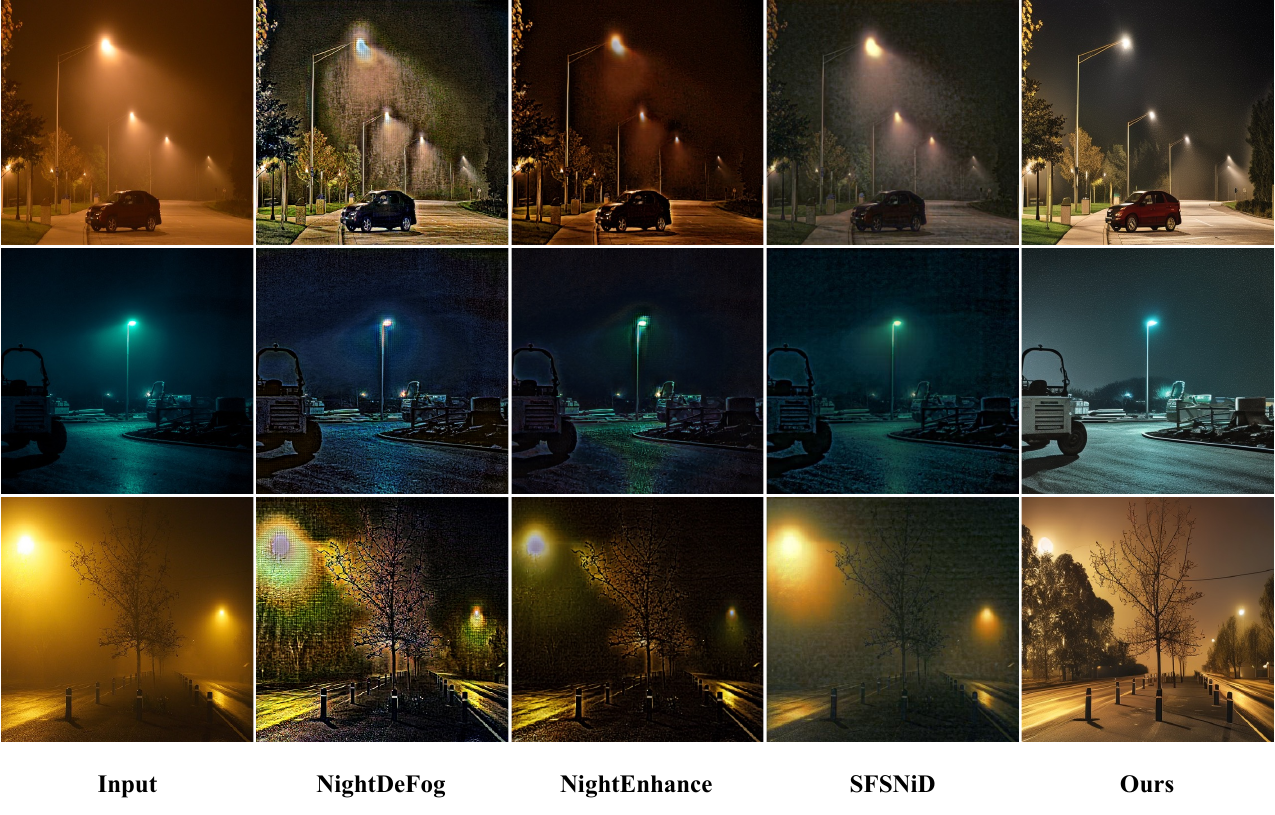}
		\vspace{-8mm}
		\captionof{figure}{Qualitative results from NightDeFog'20~\cite{yan2020nighttime}, NightEnhance'23~\cite{jin2023enhancing}, SFSNiD'24~\cite{cong2024semi} and our method on real-world data. Our method not only reduces dense haze and strong glow but also infers missing background details and content in severely degraded regions.
		}
		
		\label{fig_trailer}
	\end{center}%
}]

\begin{abstract}
Nighttime image dehazing is particularly challenging when dense haze and intense glow severely degrade or entirely obscure background information. 
Existing methods often struggle due to insufficient background priors and limited generative capability, both of which are highly important under such conditions. 
In this paper, we introduce BeyondHaze, a generative nighttime dehazing method that not only reduces haze and glow effects but also reconstructs plausible background structures in regions where visual cues are heavily degraded. 
Our approach is built on two main ideas: obtaining strong background priors by adapting image diffusion models to nighttime dehazing, and enhancing generative ability in haze- and glow-obscured areas through guided training. 
Task-specific nighttime dehazing knowledge is distilled into an image diffusion model while preserving its capacity to generate clean images.
The diffusion model is further trained on tailored image pairs to improve its ability to recover background details that are suppressed by haze effects. 
Since generative models may introduce hallucinated content, we design our framework to allow user control over the generative level, enabling a balance between visual realism and fidelity. 
Experiments on real-world nighttime images demonstrate that BeyondHaze substantially improves visibility and scene detail under dense haze.
\end{abstract}

\section{Introduction}
The degradation of nighttime imagery due to haze presents a formidable challenge, as the combination of low light conditions and atmospheric particles can significantly obscure scene details. 
Haze induces scattering of light, which not only reduces visibility but also creates glows around light sources that conceal the background. The resulting loss of visual information makes it difficult to accurately restore the original scene.

To address this problem, existing supervised nighttime dehazing methods (e.g.,~\cite{zhang2020nighttime,bernabel2024ndels,cui2023focal,cui2024omni}) rely on synthetic datasets for training. However, the considerable domain gap between real-world and synthetic degradations, especially in cases of severe haze and glow, limits the effectiveness of their learned background priors. Semi-supervised nighttime dehazing methods (e.g.,~\cite{cong2024semi, jin2023enhancing, yan2020nighttime}) combine unlabeled real data with synthetic paired data for training, but their reliance on pseudo-labels and weak supervision constrains their performance. In addition, previous methods lack the generative ability to infer background content in regions completely obscured by dense haze or intense glow. As shown in Figure \ref{fig_trailer}, such methods underperform in areas heavily degraded or entirely blocked due to haze.

In this paper, we introduce \textit{BeyondHaze}, a generative nighttime dehazing method that not only enhances the visibility of scenes affected by haze or glow but also generates plausible background details in regions where the original information is missing.
\textit{BeyondHaze} is built upon two ideas: adapting image diffusion models with nighttime dehazing knowledge to enable clean-image generation, and strengthening their generative capability to produce background content that is absent due to severe haze through guided training.

We first pretrain a dehazing model by applying haze and its associated light effects to clear nighttime images and then training the model to restore the original images. This process enables the model to learn robust background priors. After acquiring these priors, the dehazing model generates initial dehazed images and their corresponding confidence maps from real haze images. 
The before-and-after image pairs with confidence maps are then used to fine-tune a pretrained diffusion model equipped with LoRA~\cite{hu2021lora}. This approach allows our diffusion model, which possesses strong built-in world priors, to learn from real haze images and develop generative dehazing capabilities.

To enhance the model's ability to generate missing background details and occluded scene content due to haze, we introduce two additional elements into our framework: a detail enhancement model and a severe degradation model. The detail enhancement model generates training pairs by employing super-resolution on the initial dehazed images from the pretrained dehazing model. Meanwhile, the severe degradation model applies strong haze and glow effects on clear nighttime images to produce training pairs that contain substantial obscuration and occlusion. These two types of training pairs are used to fine-tune our diffusion model, enabling it to create fine-scale details and infer missing backgrounds in heavily degraded areas. 

Using these training pairs enables the model to recreate missing background information, but also introduces the risk of hallucination. Without them, results are more faithful but less visually appealing. To balance these trade-offs, our framework allows users to control the generative level through text prompts.
During inference, we also compute a generative-level map for the high-generative output, indicating how strongly each region relies on generative synthesis rather than factual restoration. This map is derived from the attention scores between tokens obtained from the high- and low-generative inference paths.

As shown in Figure \ref{fig_trailer}, our method not only reduces dense haze and strong glow but also recovers plausible background content in regions where visibility is severely degraded. In areas where the scene is fully obscured, exact recovery is unattainable, yet generating a visually coherent estimate remains useful, especially for image editing tasks. Importantly, users can control the generative level through text prompts, allowing the output to range from conservative restoration to stronger generative completion. This controllable design enables users to balance faithful recovery and visual completeness depending on the application.

Experimental results on real-world datasets show that \textit{BeyondHaze} achieves significant performance improvements. Notably, our method achieves a MUSIQ score of 65.79 and a ClipIQA score of 0.6774, surpassing existing nighttime image dehazing methods by 27.4\% in MUSIQ and 8.9\% in ClipIQA. As a summary:
\begin{itemize}
    \item 
	We introduce generative dehazing priors, where knowledge from a pretrained dehazing model is transferred to a pretrained image diffusion model. This integration equips the diffusion model with both dehazing and generative capabilities.  
    \item
    Control over the generative level of the output is made possible through the association of specific text prompts with the training pairs. 
    A generative-level map is further derived during inference to reveal potential hallucination-prone regions in the outputs.

    \item
    Both qualitative and quantitative results show that BeyondHaze significantly improves real-world nighttime haze images, effectively suppressing haze and glow while enhancing overall image quality. 
\end{itemize}

\section{Related Work}
\label{sec_related}
%
%
\paragraph{Nighttime Image Dehazing}
Existing daytime image dehazing methods~\cite{cai2016dehazenet, ren2016single, zhang2018densely, li2017aod, ren2018gated, li2018single, qu2019enhanced, ye2022perceiving, yu2022frequency, liu2019griddehazenet, dong2020multi, qin2020ffa, zheng2021ultra, wu2021contrastive, guo2022image, song2023vision, li2019semi, shao2020domain, chen2021psd, li2022physically, li2020zero, li2021you, huang2019towards, golts2019unsupervised, liu2020end, zhao2021refinednet, yang2022self, chen2024dual} can be retrained on nighttime haze datasets. However, current nighttime haze datasets are synthetic, and models trained on them struggle to handle real-world haze due to the significant domain gap between real and synthetic degradations. 

Existing nighttime image dehazing methods can be roughly divided into two categories: optimization-based and learning-based. The optimization-based dehazing methods~\cite{pei2012nighttime, ancuti2016night, ancuti2020day, zhang2014nighttime, zhang2017fast,tang2021nighttime, liu2022nighttime, wang2022variational} utilize the atmospheric scattering model or a novel imaging model to reduce nighttime haze effects. However, nighttime haze images are also influenced by other complex factors such as noise and glow, which significantly affect the accuracy of these models. 

Learning-based nighttime dehazing methods~\cite{zhang2020nighttime, kuanar2022multi, jin2023enhancing} rely on paired synthetic or unpaired real data, but the domain gap between synthetic and real degradations leads to inaccuracies on real haze images. Unsupervised and semi-supervised approaches~\cite{ jin2023enhancing, lin2024nighthaze, cong2024semi} use unpaired data or pseudo ground truths, yet unpaired training lacks pixel-level supervision and pseudo labels often fail in heavily degraded regions due to teacher-model limitations. Moreover, these methods struggle to recover regions where the true scene signal is overwhelmed by by severe haze or noise.

\paragraph{Diffusion Models for Restoration} 
Diffusion models~\cite{ho2020denoising} have shown strong performance in image restoration tasks~\cite{lin2025rgb, liu2024residual, ozdenizci2023restoring, lin2025geocomplete} such as deraining~\cite{lin2024nightrain} and raindrop removal~\cite{jin2024raindrop}. Their effectiveness, however, depends heavily on training data quality, which is a major limitation for nighttime dehazing due to the lack of real paired datasets.
Recent approaches~\cite{lan2025exploiting, tang2024realfill, parmar2024one} leverage the generative priors of pretrained diffusion models, but supervised methods~\cite{tang2024realfill, parmar2024one} remain constrained by dataset fidelity, and unpaired frameworks~\cite{lan2025exploiting} lack precise pixel-level supervision and struggle to learn strong background priors.

Our work tackles these challenges by leveraging multiple types of training data, each tailored to a specific sub-problem of nighttime dehazing. We use (1) dehazing pairs obtained by applying a SOTA dehazing model to real nighttime haze images to create pseudo ground truths without hallucinating missing backgrounds, (2) enhancement pairs generated by applying a pretrained super-resolution model to dehazed images, and (3) generative pairs produced by heavily degrading real images to simulate regions where background content would be obscured by haze.

\begin{figure*}[t!]
	\centering
    \includegraphics[width=1\linewidth]{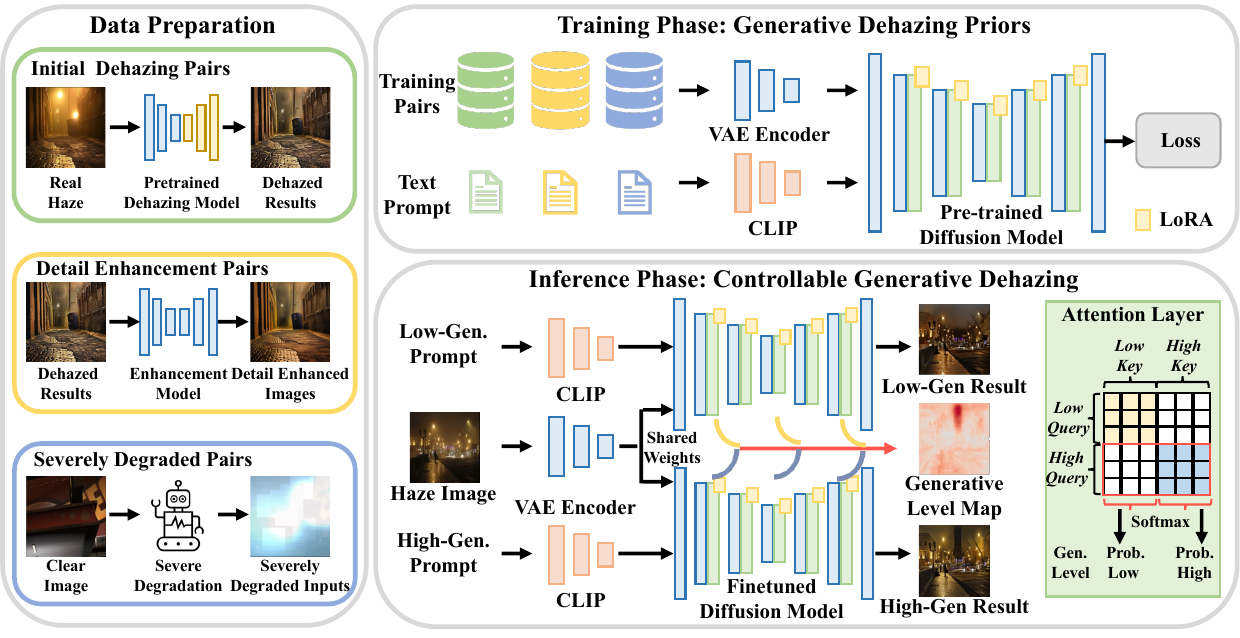}
    \caption{
    Our approach integrates dehazing priors with generative capabilities to enhance nighttime image dehazing. 
    We first pretrain a dehazing model using augmentations that simulate noise and light effects in hazy night scenes. The knowledge from this dehazing model is then distilled into a pretrained image diffusion model with LoRA. 
    Additional training pairs are generated by two supplementary models: a detail enhancement model that applies super-resolution to initial dehazed images, and a severe degradation model which produces substantially obscured images from clear images. These training pairs, combined with customized text prompts, are used to fine-tune the diffusion model and enable controllable generative dehazing. During inference, the fine-tuned model performs two inference branches conditioned on low- and high-generative text prompts, producing the corresponding outputs. A generative-level map is then derived by computing attention scores between the two branches, revealing regions that rely more on generative synthesis than on factual restoration.
  }
	\label{overview}
  \vspace{-4mm}
\end{figure*}

\section{Proposed Method: BeyondHaze}
Figure~\ref{overview} illustrates the overall pipeline of \textit{BeyondHaze}, which is based on generative dehazing priors and controllable generative dehazing. The generative dehazing priors are learned by distilling the knowledge of a pre-trained dehazing model into an image diffusion model with LoRA. Background detail and content generation are learned through training data from a detail enhancement model and a severe degradation model, where the generative level of the output is regulated by text prompts. 
During inference, the generative level of the outputs can be controlled by adjusting the text prompts. 
Furthermore, a generative-level map is obtained by calculating attention scores between tokens from the low- and high-generative inference paths, revealing potential hallucination-prone regions.

\subsection{Generative Dehazing Priors}
\label{method_spl} 
Our pre-trained dehazing model is a vision transformer \cite{he2022masked} trained with severe augmentations \cite{lin2024nighthaze}, where clear images are intentionally degraded with light effects and noise. This forces the model to learn robust background priors, enabling it to infer clear scene information despite degraded inputs.

Once the pre-trained dehazing model is obtained, we use it to generate initial dehazed images from real-world hazy nighttime images. These image pairs are then used to train our diffusion model, enabling it to progressively acquire task-specific dehazing knowledge while preserving its inherent generative properties. We refer to these integrated priors within the diffusion model as generative dehazing priors.
Optimizing a large-scale diffusion model is, however, computationally expensive and risks degrading its generative properties. To address this, we employ LoRA \cite{hu2021lora}, updating only its parameters during training. This preserves the model’s generative properties while incorporating dehazing capabilities.

Given a set of real nighttime haze images \( \{ \mathbf{x}^{\rm rh}_{i} \mid i = 1, 2, \dots, N^{\rm rh} \} \), where \( \mathbf{x}^{\rm rh}_{i} \) denotes the \( i \)-th haze image and \( N^{\rm rh} \) represents the total number of haze images, our pre-trained nighttime dehazing model, parameterized by \( w_{\rm DH} \), generates initial dehazed images, formulated as:
\begin{equation}
\label{eqn_tea}
    \mathbf{y}^{\rm rh}_{i} = f_{n}(\mathbf{x}^{\rm rh}_{i}; w_{\rm DH}),
\end{equation}
where \( \mathbf{y}^{\rm rh}_{i} \) represents the initial dehazed images for the haze image \( \mathbf{x}^{\rm rh}_{i} \), and  \( f_{n}(x; w) \) denotes the inference process. 

We additionally generate confidence maps \( \mathbf{m}^{\rm rh}_{i} \) through overlapping sampling.  
A sliding window is applied to sample overlapping regions from the input \( \mathbf{x}^{\rm rh}_{i} \), 
such that each pixel receives multiple predictions, and the variance of these predictions is used to compute the confidence map \( \mathbf{m}^{\rm rh}_{i} \).
The image pairs (i.e., real input images and their corresponding dehazed images) along with the confidence maps form a dataset \( \mathbf{D_{\rm ID}} = \{ (\mathbf{x}^{\rm rh}_{i}, \mathbf{y}^{\rm rh}_{i}, \mathbf{m}^{\rm rh}_{i}) \}_{i=1}^{N^{\rm rh}} \), which is used to fine-tune our diffusion model.

To learn dehazing priors while preserving the generative capabilities of the pre-trained diffusion model, we add a LoRA module, \( w_{\rm LoRA} \), to the diffusion model parameterized by \( w_{\rm sd} \). During training, \( w_{\rm sd} \) remains fixed, and only \( w_{\rm LoRA} \) is updated.  
We also inject a customized prompt \( t_{\rm ID} \) during training to indicate that the training pairs represent initial dehazed images generated by our dehazing model.  
The diffusion model loss for these training pairs is formulated as:
\begin{equation}
\mathcal{L}_{\text{id}} = \frac{1}{B} \sum_{j=1}^{B} \mathbb{E}_{t, \epsilon} \left[ \left\| \mathbf{m}^{\rm rh}_{j} \cdot (\epsilon - \epsilon_{\theta}(\mathbf{y}^{\rm rh}_{j}(t), t, t_{\rm {ID}}, \mathbf{x}^{\rm rh}_{j})) \right\|_2^2 \right],
\end{equation}
where \( \mathcal{L}_{\text{id}} \) denotes the diffusion loss, and \( B \) is the batch size. 
Here, \( \epsilon_{\theta}(\mathbf{y}^{\rm rh}_{j}(t), t, t_{\rm ID}, \mathbf{x}^{\rm rh}_{j}) \) is the predicted noise from the diffusion model with LoRA, parameterized by \( w_{\rm sd} \) and \( w_{\rm LoRA} \), respectively. \( \mathbf{y}^{\rm rh}_{j}(t) \) is the target image at time step \( t \), while \( t \), \( t_{\rm ID} \), and \( \mathbf{x}^{\rm rh}_{j} \) represent the conditioning inputs for the pre-trained diffusion model. 
During training, the target image \( \mathbf{y}^{\rm rh}_{j} \) and conditioning inputs \( \mathbf{x}^{\rm rh}_{j} \) are encoded into latent features via a VAE encoder. For simplicity, this step is omitted in our equation.

Fine-tuning our diffusion model on the initial dehazing pairs enables it to acquire dehazing capabilities while preserving its generative capacity through the use of LoRA.  
During inference, the fine-tuned diffusion model, guided by a text prompt, enhances the visibility of real hazy images.

\subsection{Controllable Generative Dehazing}
\label{method_GA} 
Our diffusion model with generative dehazing priors is designed to remove the visual appearance of haze and refine regions prone to artifacts. To strengthen its ability to reconstruct background details lost due to the lower signal-to-noise ratio from low light conditions and haze, and to synthesize background regions missing due to strong degradation and glows, we incorporate a detail enhancement model and a severe degradation model to produce training pairs, as depicted in Figure~\ref{overview}. Through certain prompts, the generative level of the model can be controlled by the user.

\vspace{-3mm}
\paragraph{Detail Enhancement Model}
The detail enhancement model employs image super-resolution to improve the fine-scale visual quality of the initial dehazed images generated by the pre-trained dehazing model. Since generative super-resolution methods are susceptible to hallucinating details, we include a customized text prompt together with the training pairs to enable user control. 

Given a lower-quality initial dehazed image \( \mathbf{y}^{\mathrm{rh}}_{i} \), the detail enhancement model produces a higher-quality counterpart \( \mathbf{y}^{\mathrm{hq}}_{i} \). Processing all initial dehazed images in \( \mathbf{D}_{\mathrm{ID}} \) yields a detail enhancement dataset \( \mathbf{D}_{\mathrm{DE}} = \{ (\mathbf{x}^{\mathrm{rh}}_{i}, \mathbf{y}^{\mathrm{hq}}_{i}, \mathbf{m}^{\rm rh}_{i}) \}_{i=1}^{N^{\mathrm{rh}}} \). 
Similar to the training of generative dehazing priors, we include a text prompt \( t_{\mathrm{DE}} \) with these training pairs to signify the detail enhancement. 

The detail enhancement pairs are added to the initial dehazing pairs for diffusion model training. Incorporating the text prompts allows the model to learn generative-level attributes that can be selected by the user. The diffusion model loss for detail enhancement pairs is formulated as:
\begin{equation}
\mathcal{L}_{\text{de}} = \frac{1}{B} \sum_{j=1}^{B} \mathbb{E}_{t, \epsilon} \left[ \left\| \mathbf{m}^{\rm rh}_{j} \cdot (\epsilon - \epsilon_{\theta}(\mathbf{y}^{\rm hq}_{j}(t), t, t_{\rm DE}, \mathbf{x}^{\rm rh}_{j}) )\right\|_2^2 \right].
\end{equation}
In each training batch, \(\mathcal{L}_{\text{id}}\) and \(\mathcal{L}_{\text{de}}\) share the same real-world haze samples \(\mathbf{x}^{\rm rh}\), effectively assisting the model in learning different generative levels. 

\vspace{-3mm}
\paragraph{Severe Degradation Model}
While the training data from our detail enhancement model can help the diffusion model learn to synthesize fine-scale details missing from the initial dehazed images, it does not help with recovering background regions occluded by dense haze or strong glow.  
To address this, we introduce a severe degradation model to produce training data for improving such generative capabilities in diffusion models. The degradation model severely degrades clear nighttime images by blending light effects and adding noise, creating a heavily augmented dataset, formulated as in \cite{lin2024nighthaze}:  
\begin{equation}
    \label{eqn_generate_model}
    I = W_{b}*J + (1-W_{b})*L + \epsilon,
\end{equation}
where $I$ is the augmented image, $J$ is the clear image, $W_{b}$ is the blend weight map, $L$ is the light map and  $\epsilon$ is the noise. We define the severity of the augmentation \( S \) as: 
\begin{equation}
    S = 1 - \mathbb{E} [\frac{ W_b \cdot J }{ I }]
\end{equation}
where \( \mathbb{E}[\cdot] \) represents the mean over all spatial dimensions (height, width, and channels) and $S$ defines the severity ratio.
At high severity levels, strong glow or noise significantly obscures certain regions, simulating real-world scenarios where background details are heavily degraded or entirely lost.  
Training on such data compels our diffusion model to infer missing background content from severe augmentations, enhancing its generative capability for highly degraded areas.
Additionally, it aids the diffusion model in learning the distribution of clear nighttime images, reducing potential artifacts introduced by the initial dehazing and detail enhancement training pairs.

Given clear nighttime images, we create a severe degradation dataset, defined as  
\(
\mathbf{D}_{\mathrm{BS}} = \{ (\mathbf{x}^{\mathrm{rc}}_{i}, \mathbf{y}^{\mathrm{rc}}_{i}) \}_{i=1}^{N^{\mathrm{rc}}}
\)
where \( \mathbf{x}^{\mathrm{rc}}_{i} \) and \( \mathbf{y}^{\mathrm{rc}}_{i} \) denote the \( i \)-th augmented and clear images, respectively, and \( N^{\mathrm{rc}} \) represents the total number of paired samples.  
Like for the detail enhancement model, these pairs are accompanied by a customized prompt \( t_{\mathrm{BS}} \) and added to the training set. The training loss is expressed as:  
\begin{equation}
\mathcal{L}_{\text{bs}} = \frac{1}{B} \sum_{j=1}^{B} \mathbb{E}_{t, \epsilon} \left[ \left\| \epsilon - \epsilon_{w_{\rm stu^{*}}}(\mathbf{y}^{\rm rc}_{j}(t), t, t_{\rm BS}, \mathbf{x}^{\rm rc}_{j}) \right\|_2^2 \right],
\end{equation}
where \( \mathcal{L}_{\text{bs}} \) denotes the background synthesis loss for the severe degradation data.

\vspace{-3mm}
\paragraph{Final Loss}
The final loss function is given by:
\begin{equation}
\mathcal{L}_{\text{all}} = \mathcal{L}_{\text{id}} + \mathcal{L}_{\text{de}} + \mathcal{L}_{\text{bs}}.
\end{equation}
In summary, we distill the dehazing knowledge of a pretrained dehazing model into a pretrained diffusion model using the initial dehazing loss \(\mathcal{L}_{\text{id}}\). By finetuning only the LoRA parameters while freezing the rest of the diffusion model, it retains its generative capabilities while learning to dehaze. Furthermore, the diffusion model is trained to generate fine-scale details and missing background regions via the detail enhancement loss \(\mathcal{L}_{\text{de}}\) and background synthesis loss \(\mathcal{L}_{\text{bs}}\). By associating specific prompts with the three types of training data, the user is able to choose whether or not to generate fine-scale details or background regions in the results.

\subsection{Inference with Generative-Level Map}
\label{sec:inference_level_map}

During inference, our fine-tuned diffusion model can produce dehazed results with varying generative levels by using controllable text prompts. 
Specifically, we define two types of prompts: a \textit{low-generative} prompt \(t_{\text{low}}\), which prioritizes faithful restoration with higher factual accuracy but lower perceptual richness, and a \textit{high-generative} prompt \(t_{\text{high}}\), which encourages the synthesis of missing details and visually enhanced content. 
Given an input hazy image \(\mathbf{x}\), the model performs two inference branches conditioned on these prompts, producing two corresponding outputs that represent the low- and high-generative results, respectively.

To quantify how much each region in the generative output relies on synthesis, we analyze the relationship between tokens from the two branches. In the \(\ell\)-th attention layer, the tokens from the high- and low-generative branches are concatenated along the token dimension and then projected into query and key embeddings. 
We define the merged representations as \(\mathbf{Q}_{\text{merge}}^{(\ell)} = [\mathbf{Q}_{\text{high}}^{(\ell)}; \mathbf{Q}_{\text{low}}^{(\ell)}]\) and \(\mathbf{K}_{\text{merge}}^{(\ell)} = [\mathbf{K}_{\text{high}}^{(\ell)}; \mathbf{K}_{\text{low}}^{(\ell)}]\), where \([\cdot ; \cdot]\) denotes concatenation along the token dimension. 
The attention is then computed as
\begin{equation}
\mathbf{A}^{(\ell)} = \mathrm{Softmax}\!\left(\frac{\mathbf{Q}_{\text{merge}}^{(\ell)} \left(\mathbf{K}_{\text{merge}}^{(\ell)}\right)^{\top}}{\sqrt{d}}\right),
\end{equation}
where \(d\) is the feature dimension. 

Among the merged attention matrix \(\mathbf{A}^{(\ell)} \in \mathbb{R}^{2N \times 2N}\), 
where \(N\) denotes the number of tokens in each branch, 
the submatrix corresponding to the high-generative queries \(\mathbf{Q}_{\text{high}}^{(\ell)} \in \mathbb{R}^{N \times d}\) can be written as
\[
\mathbf{A}^{(\ell)}_{\text{high}} = 
[\mathbf{A}^{(\ell)}(\mathbf{Q}_{\text{high}}^{(\ell)}, \mathbf{K}_{\text{high}}^{(\ell)}) \; ; \; 
\mathbf{A}^{(\ell)}(\mathbf{Q}_{\text{high}}^{(\ell)}, \mathbf{K}_{\text{low}}^{(\ell)})] 
\in \mathbb{R}^{N \times 2N},
\]
where each row represents the attention probability distribution of a high-generative token over the high- and low-generative keys. 
These probabilities indicate how likely each token relies on generative or restorative information. 
A higher probability toward \(\mathbf{K}_{\text{high}}^{(\ell)}\) and a lower probability toward \(\mathbf{K}_{\text{low}}^{(\ell)}\) suggest that the region is more generative, while the opposite indicates a more restoration-dominated area.

The final generative and restoration scores are obtained by averaging the layer-wise attention probabilities across all \(L\) layers:
\begin{eqnarray}
s_{\text{gen}} &=&
\frac{1}{L} \sum_{\ell=1}^{L}
\sum_{q \in \mathbf{K}_{\text{high}}^{(\ell)}}
\mathbf{A}^{(\ell)}(\mathbf{Q}_{\text{high}}^{(\ell)}, q), \\[-2pt]
s_{\text{res}} &=&
\frac{1}{L} \sum_{\ell=1}^{L}
\sum_{q \in \mathbf{K}_{\text{low}}^{(\ell)}}
\mathbf{A}^{(\ell)}(\mathbf{Q}_{\text{high}}^{(\ell)}, q).
\end{eqnarray}
where \(s_{\text{gen}}, s_{\text{res}} \in \mathbb{R}^{N \times 1}\) 
represent the averaged probabilities of attending to the high- and low-generative keys, respectively. For visualization, these scores are resized to the original image resolution to obtain the final generative-level map, which reflects the spatial distribution of generative strength across the image.
The attention computation for the generative-level map is performed solely for score extraction and does not affect the actual inference results. To ensure that the two inference branches remain independent, an attention mask is applied to prevent any information exchange between the low- and high-generative tokens.

\begin{table*}[t!]
  \centering
  \caption{Quantitative comparison on the RealNightHaze dataset (evaluated at 512$\times$512 resolution). Five non-reference metrics, namely MUSIQ~\cite{ke2021musiq}, TRES~\cite{golestaneh2022no}, ClipIQA~\cite{wang2023exploring}, HyperIQA~\cite{su2020blindly}, and ManIQA~\cite{yang2022maniqa}, are used for evaluation. For all five metrics, higher values are better. NightDeFog~\cite{yan2020nighttime}, NightEnhance~\cite{jin2023enhancing}, Diff-Dehazer \cite{lan2025exploiting} and SFSNiD~\cite{cong2024semi} are image dehazing methods, while the remaining methods are image restoration backbones. ``w/o Gen" denotes that neither generative detail enhancement nor region synthesis are selected.
  }
  \vspace{-3mm}
  \renewcommand{\arraystretch}{1.0}
  \resizebox{0.98\textwidth}{!}{
    \begin{tabular}{c|c|c|c|c|c|c|c|c|c|c}
    \toprule
    \multicolumn{1}{c|}{\textbf{Metrics}} & Uformer & Restormer & DiT   & RealFill & NightDeFog & NightEnhance & Diff-Dehazer & SFSNiD & \textbf{Ours w/o Gen} & \textbf{Ours} \\
    \midrule
    \textbf{MUSIQ $\uparrow$} & 44.33  & 45.04  & 46.02  & 51.73  & 51.61  & 50.56  & 44.57  & 43.28  & 60.29  & \textbf{65.79} \\
    \midrule
    \textbf{TRES $\uparrow$} & 51.58  & 51.89  & 53.87  & 50.25 & 68.88  & 59.08  & 50.59 & 38.50  & 68.05  & \textbf{80.08} \\
    \midrule
    \textbf{ClipIQA $\uparrow$} & 0.3256  & 0.3762  & 0.4316  & 0.4771  & 0.3743  & 0.3673  & 0.4136  & 0.3672  & 0.6217  & \textbf{0.6774} \\
    \midrule
    \textbf{HyperIQA $\uparrow$} & 0.4035  & 0.4068  & 0.4122  & 0.3556 & 0.4462  & 0.3981  & 0.3939 & 0.2839  & 0.4831 & \textbf{0.6139} \\
    \midrule
    \textbf{ManIQA $\uparrow$} & 0.3050  & 0.3210  & 0.3311  &  0.2656   & 0.3273  & 0.2879  &   0.3055  & 0.2468  & 0.3690  & \textbf{0.4964} \\
    \bottomrule
    \end{tabular}%
}
\vspace{-2mm}
   \label{tab_quan}%
\end{table*}%

\section{Experiments}
In our experiments, we test on a real-world nighttime dataset and use five no-reference image quality assessment metrics, namely MUSIQ~\cite{ke2021musiq}, ManIQA~\cite{yang2022maniqa}, ClipIQA~\cite{wang2023exploring}, HyperIQA~\cite{su2020blindly}, and TRES~\cite{golestaneh2022no}, to calculate the quality scores of the dehazed results. 

\noindent \textbf{RealNightHaze} is a real-world nighttime haze dataset consisting of 440 images. These images are sourced from previous studies \cite{zhang2020nighttime,lin2024nighthaze}. 
Due to the complex degradation factors in RealNightHaze, including haze, glow effects, extreme darkness, and noise, this dataset provides a comprehensive basis for evaluating the performance of existing methods.

\subsection{Implementation Details}
\label{sec_impl}

\noindent \textbf{Generative Dehazing Priors} 
Our generative dehazing priors are obtained from a pre-trained dehazing model and a pre-trained diffusion model. In this paper, the dehazing model is a transformer network \cite{he2022masked} pre-trained with strong augmentations \cite{lin2024nighthaze}, while \textit{Stable Diffusion v2} serves as the pre-trained diffusion model. To preserve the generative capability of the pre-trained diffusion model, we integrate a LoRA module \cite{ruiz2023dreambooth}. The rank of LoRA is set to 8.  


\begin{figure*}[t!]
	\centering
    \includegraphics[width=1\linewidth]{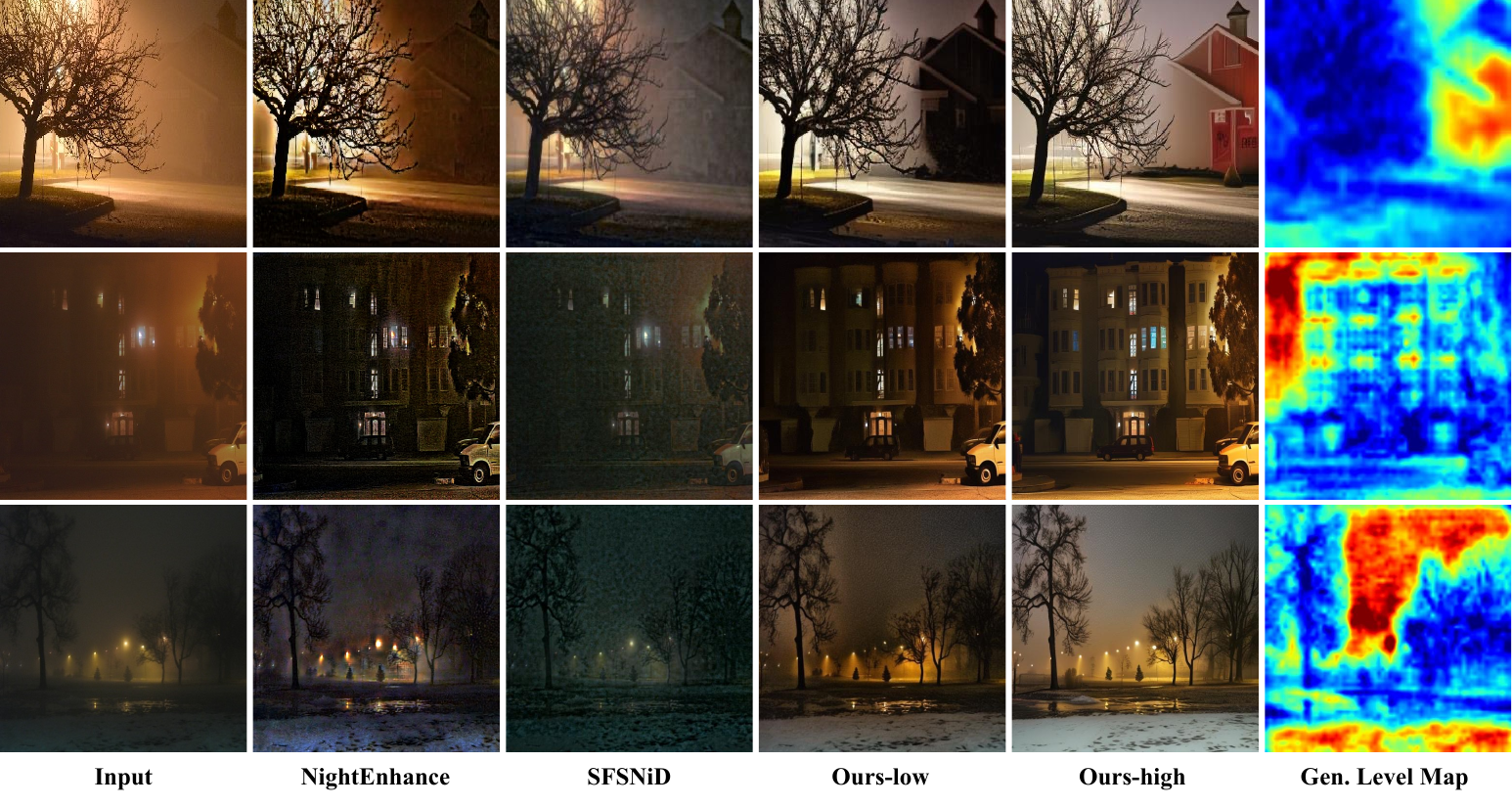}
    \vspace{-8mm}
	\caption{Qualitative results from NightEnhance’23~\cite{jin2023enhancing}, SFSNiD’24~\cite{cong2024semi}, and our method on real-world datasets. “Ours-low” refers to our low-generative results, which prioritize haze removal, while “Ours-high” refers to our high-generative results, allowing the network to infer details and backgrounds in severely degraded regions. “Gen. Level Map” shows the generative-level maps of “Ours-high,” with red indicating high and blue indicating low generative regions. Zoom in for better visualization.
	}
	\label{figs_exp1}
    \vspace{-4mm}
\end{figure*}

\begin{figure*}[t!]
	\centering
    \includegraphics[width=1\linewidth] 
    {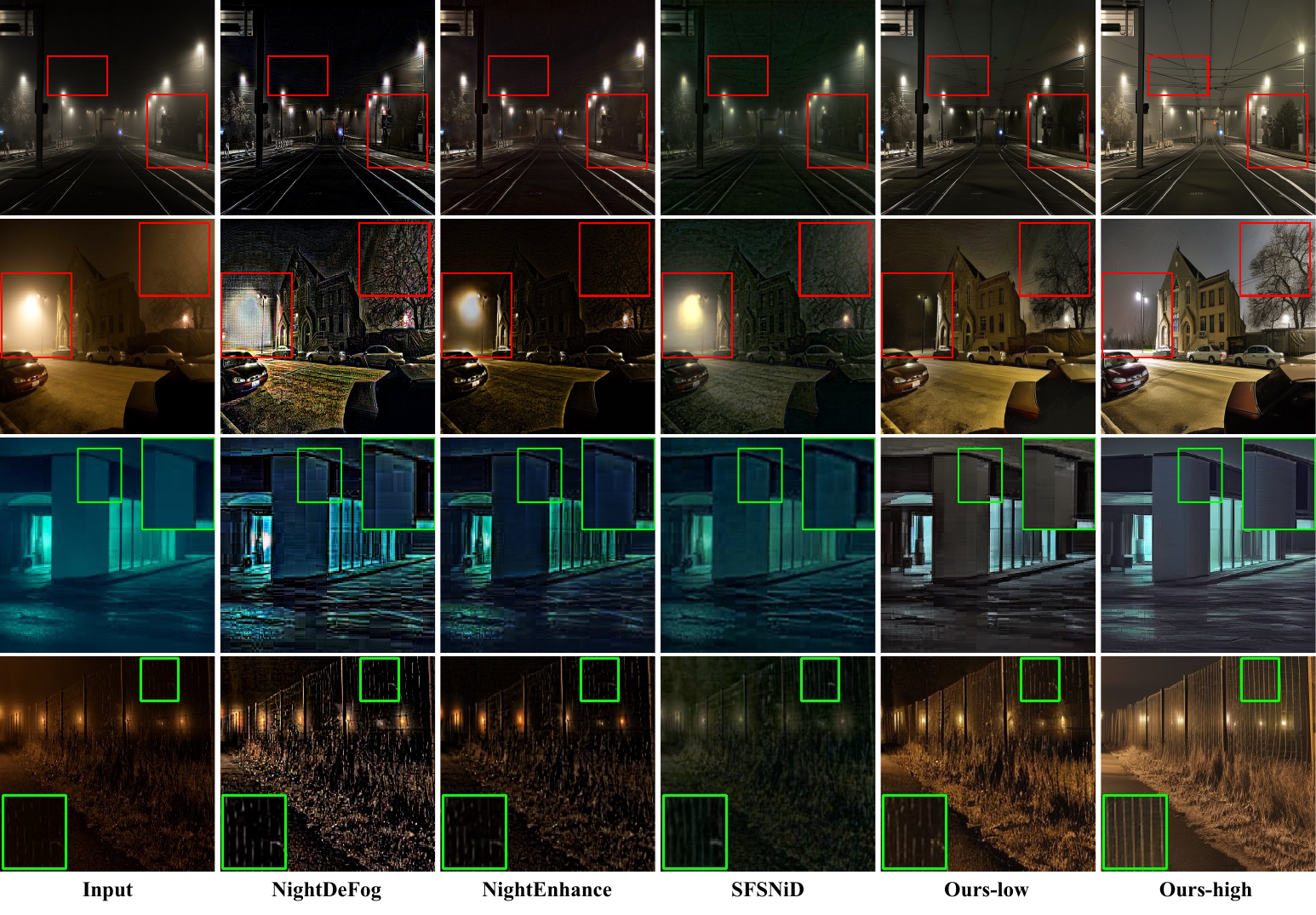}
    \vspace{-7mm}
	\caption{Qualitative results from NightDeFog’20~\cite{yan2020nighttime}, NightEnhance’23~\cite{jin2023enhancing}, SFSNiD’24~\cite{cong2024semi}, and our method on real-world datasets. “Ours-low” refers to our low-generative results, which prioritize haze removal, while “Ours-high” refers to our high-generative results, allowing the network to infer details and backgrounds in severely degraded regions. The first two rows exhibit exhibits the ability of “Ours-high” to infer missing scene content, while the last two rows display fine-scale detail generation by Ours-high. 
	}
	\label{figs_abl}
    \vspace{-5mm}
\end{figure*}

\noindent \textbf{Controllable Generative Dehazing} 
The generative capabilities of our diffusion model are enhanced by using training pairs from the detail enhancement~\cite{flux1} and severe degradation models. Additional implementation and training details are provided in the supplementary material.
During inference, given a real-world nighttime haze image, the text prompts \(t_{\text{low}}\) and \(t_{\text{high}}\) enable users to control the generative level of the outputs. By processing both prompts, our model also derives a generative-level map that highlights hallucination-prone regions in the high-generative outputs. Further details, including prompt design and inference settings, are provided in the supplementary material.

\subsection{Quantitative Evaluation on RealNightHaze}

We evaluate the performance of our \textit{BeyondHaze} method on the RealNightHaze dataset, with experimental results presented in Table~\ref{tab_quan}. We compare our method against several state-of-the-art techniques, including Uformer~\cite{wang2022uformer}, Restormer~\cite{zamir2022restormer}, DiT~\cite{Peebles2023DiT}, RealFill~\cite{tang2024realfill}, NightDeFog~\cite{yan2020nighttime}, NightEnhance~\cite{jin2023enhancing}, Diff-Dehazer~\cite{lan2025exploiting} and SFSNiD~\cite{cong2024semi}. Uformer, Restormer, DiT and RealFill are restoration backbones trained on synthetic datasets, while the remaining methods are specifically designed for image dehazing. 


Table~\ref{tab_quan} shows that \textit{BeyondHaze} consistently improves all non-reference metrics. Synthetic-trained models such as Restormer and DiT reach only ~45 MUSIQ, while nighttime dehazing methods like NightDeFog and NightEnhance perform better (e.g., NightDeFog: 51.61 MUSIQ, 68.88 TRES). Our method further increases these scores to 65.79 MUSIQ and 80.08 TRES.

\subsection{Qualitative Evaluation on RealNightHaze} 
Figure~\ref{figs_exp1} presents a qualitative comparison of different methods on real-world nighttime haze data. Existing nighttime dehazing approaches often suffer from noise artifacts, produce dark outputs, and fail to recover background content in severely degraded regions (e.g., dense haze or strong glow). As shown in Figures~\ref{figs_exp1} and~\ref{figs_abl}, they tend to generate black areas in such cases. The main reason is their lack of generative capability, which prevents them from inferring the missing background content. 
In contrast, our \textit{BeyondHaze} leverages strong priors from a pre-trained diffusion model to not only address haze effects but also infer backgrounds in regions entirely occluded by dense haze. As shown in Figures \ref{figs_exp1} and \ref{figs_abl}, our method achieves significant performance improvements. Moreover, we introduce a generative-level map during inference to visualize regions prone to generative synthesis or hallucination. Fig.~\ref{figs_exp1} shows that the red-highlighted areas in this map indicate regions likely to contain hallucinated content, such as the house color and texture, building boundaries, and sky regions.

\subsection{Ablation Studies}

In this section, we perform ablation studies to validate our generative dehazing priors and controllable generative dehazing. The experimental results are presented in Table \ref{tab_ablation}. 

\noindent \textbf{Generative Dehazing Priors} Our pre-trained dehazing model achieves a MUSIQ score of 58.92, a TRES score of 63.83, and a ClipIQA score of 0.5290. In contrast, our finetuned diffusion model achieves a MUSIQ score of 60.29 and a ClipIQA score of 0.6217, clearly surpassing the pre-trained dehazing model. This demonstrates that distilling knowledge from a pre-trained dehazing model to an image diffusion model leads to higher performance than the dehazing teacher model.



\begin{table}[t]
  \centering

  \caption{
  Ablation studies on real-world haze datasets. ``GDP" denotes our generative dehazing priors. We use three non-reference metrics for evaluation: MUSIQ~\cite{ke2021musiq}, TRES~\cite{golestaneh2022no}, and ClipIQA~\cite{wang2023exploring}. The scores in the first row are calculated from the results produced by our pre-trained dehazing model.
  }
  \vspace{-3mm}
  \renewcommand{\arraystretch}{1.0}
  \resizebox{0.48\textwidth}{!}{
    \begin{tabular}{c|c|c|c|c|c}
    \toprule
    \multirow{2}[2]{*}{GDP} & \multicolumn{2}{c|}{Supplementary Models} & \multirow{2}[2]{*}{\textbf{MUSIQ $\uparrow$}} & \multirow{2}[2]{*}{\textbf{TRES $\uparrow$}} & \multirow{2}[2]{*}{\textbf{ClipIQA $\uparrow$}} \\
\cmidrule{2-3}          & Detail & Degradation &       &       &  \\
    \midrule
       $\times$   &    $\times$    &    $\times$    & 58.92 &	63.83 &	0.5290   \\
    \midrule
    \( \checkmark\)     &    $\times$    &     $\times$   & 60.29 &	68.05 &	0.6217   \\
    \midrule
     \( \checkmark\)      &   \( \checkmark\)      &   $\times$     &   63.95 &	79.91 &	0.6570   \\
    \midrule
     \( \checkmark\)      &   \( \checkmark\)      &  \( \checkmark\)      & \textbf{65.79} &	\textbf{80.08} &	\textbf{0.6774}  \\
    \bottomrule
    \end{tabular}%
    }
    \vspace{-3mm}
  \label{tab_ablation}%
\end{table}%

\noindent \textbf{Controllable Generative Dehazing} Table \ref{tab_ablation} shows that our supplementary models designed for generative details and backgrounds lead to a substantial performance improvement. Including them yields a MUSIQ score of 65.79, outperforming the pre-trained dehazing model by 6.87 points. This improvement demonstrates that the two models are effective at elevating the generative capabilities of the pre-trained diffusion model, resulting in superior performance. 

\noindent \textbf{Different Generative Levels} Using text prompts, our fine-tuned diffusion model can control the generative level of the outputs. The experimental results are shown in Figure \ref{figs_abl}. Our low-generative results, which prioritize haze removal, effectively mitigate haze effects but fail to infer background details and content in severely degraded regions. In contrast, our high-generative results enable the network to infer background details and content in these regions, significantly boosting visual quality.

\section{Conclusion} 
This paper presents \textit{BeyondHaze}, a novel generative approach for nighttime image dehazing that integrates dehazing priors with generative capabilities. 
We first pre-train a dehazing model using strong noise and glow augmentations to learn robust priors, which are then distilled into a diffusion model while preserving its generative power.
%
The generative power of the diffusion model to recover fine-scale details and missing background regions in nighttime haze images is enhanced by training data from two auxiliary models: 
a detail enhancement model and a severe degradation model. The detail enhancement model refines dehazed regions with lost details to produce detail enhancement training pairs, while the severe degradation model creates training pairs from clear images with severe augmentations to aid the diffusion model in learning to synthesizes missing background areas. Customized text prompts are associated with these training pairs to allow users to control the level of generative dehazing in the outputs. A generative-level map is further derived during inference to indicate regions that rely more on generative synthesis than on factual restoration.
Extensive experiments on real-world nighttime datasets demonstrate that \textit{BeyondHaze} achieves substantial improvements over other methods.

{
    \small
    \bibliographystyle{ieeenat_fullname}
    \bibliography{main}

@String(CVPR= {IEEE Conf. Comput. Vis. Pattern Recog.})

@String(ICCV= {Int. Conf. Comput. Vis.})

@String(TOG= {ACM Trans. Graph.})

@String(ICIP = {IEEE Int. Conf. Image Process.})

@String(AAAI = {AAAI})

@String(CVPR  = {CVPR})

@String(ICCV  = {ICCV})

@String(TOG   = {ACM TOG})

@String(ICIP  = {ICIP})

@inproceedings{ye2022perceiving,
	title={Perceiving and Modeling Density for Image Dehazing},
	author={Ye, Tian and Zhang, Yunchen and Jiang, Mingchao and Chen, Liang and Liu, Yun and Chen, Sixiang and Chen, Erkang},
	booktitle={Computer Vision--ECCV 2022: 17th European Conference, Tel Aviv, Israel, October 23--27, 2022, Proceedings, Part XIX},
	pages={130--145},
	year={2022},
	organization={Springer}
}

@inproceedings{yu2022frequency,
	title={Frequency and spatial dual guidance for image dehazing},
	author={Yu, Hu and Zheng, Naishan and Zhou, Man and Huang, Jie and Xiao, Zeyu and Zhao, Feng},
	booktitle={Computer Vision--ECCV 2022: 17th European Conference, Tel Aviv, Israel, October 23--27, 2022, Proceedings, Part XIX},
	pages={181--198},
	year={2022},
	organization={Springer}
}

@article{wang2022variational,
	title={Variational Single Nighttime Image Haze Removal With a Gray Haze-Line Prior},
	author={Wang, Wenhui and Wang, Anna and Liu, Chen},
	journal={IEEE Transactions on Image Processing},
	volume={31},
	pages={1349--1363},
	year={2022},
	publisher={IEEE}
}

@article{tang2021nighttime,
	title={Nighttime image dehazing based on Retinex and dark channel prior using Taylor series expansion},
	author={Tang, Qunfang and Yang, Jie and He, Xiangjian and Jia, Wenjing and Zhang, Qingnian and Liu, Haibo},
	journal={Computer Vision and Image Understanding},
	volume={202},
	pages={103086},
	year={2021},
	publisher={Elsevier}
}

@article{liu2020end,
	title={End-to-end single image fog removal using enhanced cycle consistent adversarial networks},
	author={Liu, Wei and Hou, Xianxu and Duan, Jiang and Qiu, Guoping},
	journal={IEEE Transactions on Image Processing},
	volume={29},
	pages={7819--7833},
	year={2020},
	publisher={IEEE}
}

@article{li2020zero,
	title={Zero-shot image dehazing},
	author={Li, Boyun and Gou, Yuanbiao and Liu, Jerry Zitao and Zhu, Hongyuan and Zhou, Joey Tianyi and Peng, Xi},
	journal={IEEE Transactions on Image Processing},
	volume={29},
	pages={8457--8466},
	year={2020},
	publisher={IEEE}
}

@article{zhao2021refinednet,
	title={RefineDNet: A weakly supervised refinement framework for single image dehazing},
	author={Zhao, Shiyu and Zhang, Lin and Shen, Ying and Zhou, Yicong},
	journal={IEEE Transactions on Image Processing},
	volume={30},
	pages={3391--3404},
	year={2021},
	publisher={IEEE}
}

@article{li2021you,
	title={You only look yourself: Unsupervised and untrained single image dehazing neural network},
	author={Li, Boyun and Gou, Yuanbiao and Gu, Shuhang and Liu, Jerry Zitao and Zhou, Joey Tianyi and Peng, Xi},
	journal={International Journal of Computer Vision},
	volume={129},
	number={5},
	pages={1754--1767},
	year={2021},
	publisher={Springer}
}

@inproceedings{wu2021contrastive,
	title={Contrastive learning for compact single image dehazing},
	author={Wu, Haiyan and Qu, Yanyun and Lin, Shaohui and Zhou, Jian and Qiao, Ruizhi and Zhang, Zhizhong and Xie, Yuan and Ma, Lizhuang},
	booktitle={Proceedings of the IEEE/CVF Conference on Computer Vision and Pattern Recognition},
	pages={10551--10560},
	year={2021}
}

@inproceedings{li2022physically,
	title={Physically Disentangled Intra-and Inter-Domain Adaptation for Varicolored Haze Removal},
	author={Li, Yi and Chang, Yi and Gao, Yan and Yu, Changfeng and Yan, Luxin},
	booktitle={Proceedings of the IEEE/CVF Conference on Computer Vision and Pattern Recognition},
	pages={5841--5850},
	year={2022}
}

@article{song2023vision,
	title={Vision transformers for single image dehazing},
	author={Song, Yuda and He, Zhuqing and Qian, Hui and Du, Xin},
	journal={IEEE Transactions on Image Processing},
	volume={32},
	pages={1927--1941},
	year={2023},
	publisher={IEEE}
}

@inproceedings{yang2022self,
	title={Self-Augmented Unpaired Image Dehazing via Density and Depth Decomposition},
	author={Yang, Yang and Wang, Chaoyue and Liu, Risheng and Zhang, Lin and Guo, Xiaojie and Tao, Dacheng},
	booktitle={Proceedings of the IEEE/CVF Conference on Computer Vision and Pattern Recognition},
	pages={2037--2046},
	year={2022}
}

@inproceedings{guo2022image,
	title={Image Dehazing Transformer with Transmission-Aware 3D Position Embedding},
	author={Guo, Chun-Le and Yan, Qixin and Anwar, Saeed and Cong, Runmin and Ren, Wenqi and Li, Chongyi},
	booktitle={Proceedings of the IEEE/CVF Conference on Computer Vision and Pattern Recognition},
	pages={5812--5820},
	year={2022}
}

@article{li2019semi,
	title={Semi-supervised image dehazing},
	author={Li, Lerenhan and Dong, Yunlong and Ren, Wenqi and Pan, Jinshan and Gao, Changxin and Sang, Nong and Yang, Ming-Hsuan},
	journal={IEEE Transactions on Image Processing},
	volume={29},
	pages={2766--2779},
	year={2019},
	publisher={IEEE}
}

@inproceedings{zhang2018densely,
	title={Densely connected pyramid dehazing network},
	author={Zhang, He and Patel, Vishal M},
	booktitle={Proceedings of the IEEE conference on computer vision and pattern recognition},
	pages={3194--3203},
	year={2018}
}

@inproceedings{li2018single,
	title={Single image dehazing via conditional generative adversarial network},
	author={Li, Runde and Pan, Jinshan and Li, Zechao and Tang, Jinhui},
	booktitle={Proceedings of the IEEE Conference on Computer Vision and Pattern Recognition},
	pages={8202--8211},
	year={2018}
}

@inproceedings{ren2018gated,
	title={Gated fusion network for single image dehazing},
	author={Ren, Wenqi and Ma, Lin and Zhang, Jiawei and Pan, Jinshan and Cao, Xiaochun and Liu, Wei and Yang, Ming-Hsuan},
	booktitle={Proceedings of the IEEE conference on computer vision and pattern recognition},
	pages={3253--3261},
	year={2018}
}

@inproceedings{ren2016single,
	title={Single image dehazing via multi-scale convolutional neural networks},
	author={Ren, Wenqi and Liu, Si and Zhang, Hua and Pan, Jinshan and Cao, Xiaochun and Yang, Ming-Hsuan},
	booktitle={European conference on computer vision},
	pages={154--169},
	year={2016},
	organization={Springer}
}

@inproceedings{li2017aod,
	title={Aod-net: All-in-one dehazing network},
	author={Li, Boyi and Peng, Xiulian and Wang, Zhangyang and Xu, Jizheng and Feng, Dan},
	booktitle={Proceedings of the IEEE international conference on computer vision},
	pages={4770--4778},
	year={2017}
}

@article{cai2016dehazenet,
	title={Dehazenet: An end-to-end system for single image haze removal},
	author={Cai, Bolun and Xu, Xiangmin and Jia, Kui and Qing, Chunmei and Tao, Dacheng},
	journal={IEEE Transactions on Image Processing},
	volume={25},
	number={11},
	pages={5187--5198},
	year={2016},
	publisher={IEEE}
}

@inproceedings{huang2019towards,
	title={Towards Unsupervised Single Image Dehazing With Deep Learning},
	author={Huang, Lu-Yao and Yin, Jia-Li and Chen, Bo-Hao and Ye, Shao-Zhen},
	booktitle={2019 IEEE International Conference on Image Processing (ICIP)},
	pages={2741--2745},
	year={2019},
	organization={IEEE}
}

@article{golts2019unsupervised,
	title={Unsupervised single image dehazing using dark channel prior loss},
	author={Golts, Alona and Freedman, Daniel and Elad, Michael},
	journal={IEEE Transactions on Image Processing},
	volume={29},
	pages={2692--2701},
	year={2019},
	publisher={IEEE}
}

@inproceedings{shao2020domain,
	title={Domain Adaptation for Image Dehazing},
	author={Shao, Yuanjie and Li, Lerenhan and Ren, Wenqi and Gao, Changxin and Sang, Nong},
	booktitle={Proceedings of the IEEE/CVF Conference on Computer Vision and Pattern Recognition},
	pages={2808--2817},
	year={2020}
}

@inproceedings{liu2022nighttime,
	title={Nighttime Image Dehazing Based on Variational Decomposition Model},
	author={Liu, Yun and Yan, Zhongsheng and Wu, Aimin and Ye, Tian and Li, Yuche},
	booktitle={Proceedings of the IEEE/CVF Conference on Computer Vision and Pattern Recognition},
	pages={640--649},
	year={2022}
}

@article{kuanar2022multi,
	title={Multi-path dilated convolution network for haze and glow removal in nighttime images},
	author={Kuanar, Shiba and Mahapatra, Dwarikanath and Bilas, Monalisa and Rao, KR},
	journal={The Visual Computer},
	volume={38},
	number={3},
	pages={1121--1134},
	year={2022},
	publisher={Springer}
}

@inproceedings{chen2021psd,
	title={PSD: Principled synthetic-to-real dehazing guided by physical priors},
	author={Chen, Zeyuan and Wang, Yangchao and Yang, Yang and Liu, Dong},
	booktitle={Proceedings of the IEEE/CVF Conference on Computer Vision and Pattern Recognition},
	pages={7180--7189},
	year={2021}
}

@inproceedings{qin2020ffa,
	title={FFA-Net: Feature fusion attention network for single image dehazing},
	author={Qin, Xu and Wang, Zhilin and Bai, Yuanchao and Xie, Xiaodong and Jia, Huizhu},
	booktitle={Proceedings of the AAAI Conference on Artificial Intelligence},
	volume={34},
	number={07},
	pages={11908--11915},
	year={2020}
}

@inproceedings{zheng2021ultra,
	title={Ultra-high-definition image dehazing via multi-guided bilateral learning},
	author={Zheng, Zhuoran and Ren, Wenqi and Cao, Xiaochun and Hu, Xiaobin and Wang, Tao and Song, Fenglong and Jia, Xiuyi},
	booktitle={2021 IEEE/CVF Conference on Computer Vision and Pattern Recognition (CVPR)},
	pages={16180--16189},
	year={2021},
	organization={IEEE}
}

@inproceedings{dong2020multi,
	title={Multi-scale boosted dehazing network with dense feature fusion},
	author={Dong, Hang and Pan, Jinshan and Xiang, Lei and Hu, Zhe and Zhang, Xinyi and Wang, Fei and Yang, Ming-Hsuan},
	booktitle={Proceedings of the IEEE/CVF conference on computer vision and pattern recognition},
	pages={2157--2167},
	year={2020}
}

@inproceedings{liu2019griddehazenet,
	title={Griddehazenet: Attention-based multi-scale network for image dehazing},
	author={Liu, Xiaohong and Ma, Yongrui and Shi, Zhihao and Chen, Jun},
	booktitle={Proceedings of the IEEE/CVF International Conference on Computer Vision},
	pages={7314--7323},
	year={2019}
}

@inproceedings{pei2012nighttime,
	title={Nighttime haze removal using color transfer pre-processing and dark channel prior},
	author={Pei, Soo-Chang and Lee, Tzu-Yen},
	booktitle={2012 19th IEEE International conference on image processing},
	pages={957--960},
	year={2012},
	organization={IEEE}
}

@inproceedings{zhang2020nighttime,
	title={Nighttime dehazing with a synthetic benchmark},
	author={Zhang, Jing and Cao, Yang and Zha, Zheng-Jun and Tao, Dacheng},
	booktitle={Proceedings of the 28th ACM International Conference on Multimedia},
	pages={2355--2363},
	year={2020}
}

@inproceedings{zhang2014nighttime,
	title={Nighttime haze removal based on a new imaging model},
	author={Zhang, Jing and Cao, Yang and Wang, Zengfu},
	booktitle=P_ICIP,
	pages={4557--4561},
	year={2014},
	organization={IEEE}
}

@inproceedings{qu2019enhanced,
	title={Enhanced Pix2pix Dehazing Network},
	author={Qu, Yanyun and Chen, Yizi and Huang, Jingying and Xie, Yuan},
	booktitle={Proceedings of the IEEE Conference on Computer Vision and Pattern Recognition},
	pages={8160--8168},
	year={2019}
}

@inproceedings{yan2020nighttime,
	title={Nighttime defogging using high-low frequency decomposition and grayscale-color networks},
	author={Yan, Wending and Tan, Robby T and Dai, Dengxin},
	booktitle={European Conference on Computer Vision},
	pages={473--488},
	year={2020},
	organization={Springer}
}

@inproceedings{zhang2017fast,
	title={Fast haze removal for nighttime image using maximum reflectance prior},
	author={Zhang, Jing and Cao, Yang and Fang, Shuai and Kang, Yu and Wen Chen, Chang},
	booktitle={Proceedings of the IEEE conference on computer vision and pattern recognition},
	pages={7418--7426},
	year={2017}
}

@inproceedings{ancuti2016night,
	title={Night-time dehazing by fusion},
	author={Ancuti, Cosmin and Ancuti, Codruta O and De Vleeschouwer, Christophe and Bovik, Alan C},
	booktitle={2016 IEEE International Conference on Image Processing (ICIP)},
	pages={2256--2260},
	year={2016},
	organization={IEEE}
}

@article{ancuti2020day,
	title={Day and night-time dehazing by local airlight estimation},
	author={Ancuti, Cosmin and Ancuti, Codruta O and De Vleeschouwer, Christophe and Bovik, Alan C},
	journal={IEEE Transactions on Image Processing},
	volume={29},
	pages={6264--6275},
	year={2020},
	publisher={IEEE}
}

@inproceedings{jin2023enhancing,
  title={Enhancing visibility in nighttime haze images using guided apsf and gradient adaptive convolution},
  author={Jin, Yeying and Lin, Beibei and Yan, Wending and Yuan, Yuan and Ye, Wei and Tan, Robby T},
  booktitle={Proceedings of the 31st ACM International Conference on Multimedia},
  pages={2446--2457},
  year={2023}
}

@inproceedings{ke2021musiq,
  title={Musiq: Multi-scale image quality transformer},
  author={Ke, Junjie and Wang, Qifei and Wang, Yilin and Milanfar, Peyman and Yang, Feng},
  booktitle={Proceedings of the IEEE/CVF International Conference on Computer Vision},
  pages={5148--5157},
  year={2021}
}

@inproceedings{yang2022maniqa,
  title={Maniqa: Multi-dimension attention network for no-reference image quality assessment},
  author={Yang, Sidi and Wu, Tianhe and Shi, Shuwei and Lao, Shanshan and Gong, Yuan and Cao, Mingdeng and Wang, Jiahao and Yang, Yujiu},
  booktitle={Proceedings of the IEEE/CVF Conference on Computer Vision and Pattern Recognition},
  pages={1191--1200},
  year={2022}
}

@inproceedings{su2020blindly,
  title={Blindly assess image quality in the wild guided by a self-adaptive hyper network},
  author={Su, Shaolin and Yan, Qingsen and Zhu, Yu and Zhang, Cheng and Ge, Xin and Sun, Jinqiu and Zhang, Yanning},
  booktitle={Proceedings of the IEEE/CVF Conference on Computer Vision and Pattern Recognition},
  pages={3667--3676},
  year={2020}
}

@inproceedings{wang2023exploring,
  title={Exploring clip for assessing the look and feel of images},
  author={Wang, Jianyi and Chan, Kelvin CK and Loy, Chen Change},
  booktitle={Proceedings of the AAAI Conference on Artificial Intelligence},
  volume={37},
  number={2},
  pages={2555--2563},
  year={2023}
}

@inproceedings{golestaneh2022no,
  title={No-reference image quality assessment via transformers, relative ranking, and self-consistency},
  author={Golestaneh, S Alireza and Dadsetan, Saba and Kitani, Kris M},
  booktitle={Proceedings of the IEEE/CVF Winter Conference on Applications of Computer Vision},
  pages={1220--1230},
  year={2022}
}

@inproceedings{he2022masked,
  title={Masked autoencoders are scalable vision learners},
  author={He, Kaiming and Chen, Xinlei and Xie, Saining and Li, Yanghao and Doll{\'a}r, Piotr and Girshick, Ross},
  booktitle={Proceedings of the IEEE/CVF conference on computer vision and pattern recognition},
  pages={16000--16009},
  year={2022}
}

@inproceedings{zamir2022restormer,
  title={Restormer: Efficient transformer for high-resolution image restoration},
  author={Zamir, Syed Waqas and Arora, Aditya and Khan, Salman and Hayat, Munawar and Khan, Fahad Shahbaz and Yang, Ming-Hsuan},
  booktitle={Proceedings of the IEEE/CVF conference on computer vision and pattern recognition},
  pages={5728--5739},
  year={2022}
}

@InProceedings{Peebles2023DiT,
    author    = {Peebles, William and Xie, Saining},
    title     = {Scalable Diffusion Models with Transformers},
    booktitle = {Proceedings of the IEEE/CVF International Conference on Computer Vision (ICCV)},
    month     = {October},
    year      = {2023},
    pages     = {4195-4205}
}

@inproceedings{Wang2022Uformer,
	title={Uformer: A general u-shaped transformer for image restoration},
	author={Wang, Zhendong and Cun, Xiaodong and Bao, Jianmin and Zhou, Wengang and Liu, Jianzhuang and Li, Houqiang},
    booktitle       = {CVPR},
	year={2022}
}

@article{lin2024nighthaze,
  title={NightHaze: Nighttime Image Dehazing via Self-Prior Learning},
  author={Lin, Beibei and Jin, Yeying and Yan, Wending and Ye, Wei and Yuan, Yuan and Tan, Robby T},
  journal={arXiv preprint arXiv:2403.07408},
  year={2024}
}

@inproceedings{cong2024semi,
  title={A Semi-supervised Nighttime Dehazing Baseline with Spatial-Frequency Aware and Realistic Brightness Constraint},
  author={Cong, Xiaofeng and Gui, Jie and Zhang, Jing and Hou, Junming and Shen, Hao},
  booktitle={Proceedings of the IEEE/CVF Conference on Computer Vision and Pattern Recognition},
  pages={2631--2640},
  year={2024}
}

@article{hu2021lora,
  title={Lora: Low-rank adaptation of large language models},
  author={Hu, Edward J and Shen, Yelong and Wallis, Phillip and Allen-Zhu, Zeyuan and Li, Yuanzhi and Wang, Shean and Wang, Lu and Chen, Weizhu},
  journal={arXiv preprint arXiv:2106.09685},
  year={2021}
}

@misc{flux1,
  author       = {JasperAI},
  title        = {Flux.1 -- FluxControlNet Upscaler},
  year         = {2024},
  version      = {1.0},
  url          = {https://github.com/jasperai/flux.1},
  note         = {Software available from https://github.com/jasperai/flux.1}
}

@inproceedings{ruiz2023dreambooth,
  title={Dreambooth: Fine tuning text-to-image diffusion models for subject-driven generation},
  author={Ruiz, Nataniel and Li, Yuanzhen and Jampani, Varun and Pritch, Yael and Rubinstein, Michael and Aberman, Kfir},
  booktitle={Proceedings of the IEEE/CVF conference on computer vision and pattern recognition},
  pages={22500--22510},
  year={2023}
}

@article{bernabel2024ndels,
  title={NDELS: A novel approach for nighttime dehazing, low-light enhancement, and light suppression},
  author={Bernabel, Silvano A and Agaian, Sos S},
  journal={IEEE Transactions on Multimedia},
  year={2024},
  publisher={IEEE}
}

@inproceedings{cui2023focal,
  title={Focal network for image restoration},
  author={Cui, Yuning and Ren, Wenqi and Cao, Xiaochun and Knoll, Alois},
  booktitle={Proceedings of the IEEE/CVF international conference on computer vision},
  pages={13001--13011},
  year={2023}
}

@inproceedings{cui2024omni,
  title={Omni-Kernel Network for Image Restoration},
  author={Cui, Yuning and Ren, Wenqi and Knoll, Alois},
  booktitle={Proceedings of the AAAI Conference on Artificial Intelligence},
  volume={38},
  number={2},
  pages={1426--1434},
  year={2024}
}

@article{ho2020denoising,
  title={Denoising diffusion probabilistic models},
  author={Ho, Jonathan and Jain, Ajay and Abbeel, Pieter},
  journal={Advances in neural information processing systems},
  volume={33},
  pages={6840--6851},
  year={2020}
}

@inproceedings{lin2024nightrain,
  title={NightRain: Nighttime Video Deraining via Adaptive-Rain-Removal and Adaptive-Correction},
  author={Lin, Beibei and Jin, Yeying and Yan, Wending and Ye, Wei and Yuan, Yuan and Zhang, Shunli and Tan, Robby T},
  booktitle={Proceedings of the AAAI Conference on Artificial Intelligence},
  volume={38},
  number={4},
  pages={3378--3385},
  year={2024}
}

@article{ozdenizci2023restoring,
  title={Restoring vision in adverse weather conditions with patch-based denoising diffusion models},
  author={{\"O}zdenizci, Ozan and Legenstein, Robert},
  journal={IEEE Transactions on Pattern Analysis and Machine Intelligence},
  volume={45},
  number={8},
  pages={10346--10357},
  year={2023},
  publisher={IEEE}
}

@inproceedings{liu2024residual,
  title={Residual denoising diffusion models},
  author={Liu, Jiawei and Wang, Qiang and Fan, Huijie and Wang, Yinong and Tang, Yandong and Qu, Liangqiong},
  booktitle={Proceedings of the IEEE/CVF Conference on Computer Vision and Pattern Recognition},
  pages={2773--2783},
  year={2024}
}

@inproceedings{jin2024raindrop,
  title={Raindrop Clarity: A Dual-Focused Dataset for Day and Night Raindrop Removal},
  author={Jin, Yeying and Li, Xin and Wang, Jiadong and Zhang, Yan and Zhang, Malu},
  booktitle={European Conference on Computer Vision},
  pages={1--17},
  year={2024},
  organization={Springer}
}

@inproceedings{wang2021real,
  title={Real-esrgan: Training real-world blind super-resolution with pure synthetic data},
  author={Wang, Xintao and Xie, Liangbin and Dong, Chao and Shan, Ying},
  booktitle={Proceedings of the IEEE/CVF international conference on computer vision},
  pages={1905--1914},
  year={2021}
}

@inproceedings{lan2025exploiting,
  title={Exploiting diffusion prior for real-world image dehazing with unpaired training},
  author={Lan, Yunwei and Cui, Zhigao and Liu, Chang and Peng, Jialun and Wang, Nian and Luo, Xin and Liu, Dong},
  booktitle={Proceedings of the AAAI Conference on Artificial Intelligence},
  volume={39},
  number={4},
  pages={4455--4463},
  year={2025}
}

@article{tang2024realfill,
  title={Realfill: Reference-driven generation for authentic image completion},
  author={Tang, Luming and Ruiz, Nataniel and Chu, Qinghao and Li, Yuanzhen and Holynski, Aleksander and Jacobs, David E and Hariharan, Bharath and Pritch, Yael and Wadhwa, Neal and Aberman, Kfir and others},
  journal={ACM Transactions on Graphics (TOG)},
  volume={43},
  number={4},
  pages={1--12},
  year={2024},
  publisher={ACM New York, NY, USA}
}

@article{parmar2024one,
  title={One-step image translation with text-to-image models},
  author={Parmar, Gaurav and Park, Taesung and Narasimhan, Srinivasa and Zhu, Jun-Yan},
  journal={arXiv preprint arXiv:2403.12036},
  year={2024}
}

@inproceedings{chen2024dual,
  title={Dual-rain: Video rain removal using assertive and gentle teachers},
  author={Chen, Tingting and Lin, Beibei and Jin, Yeying and Yan, Wending and Ye, Wei and Yuan, Yuan and Tan, Robby T},
  booktitle={European Conference on Computer Vision},
  pages={127--143},
  year={2024},
  organization={Springer}
}

@article{lin2025rgb,
  title={RGB-to-Polarization Estimation: A New Task and Benchmark Study},
  author={Lin, Beibei and Yuan, Zifeng and Chen, Tingting},
  journal={arXiv preprint arXiv:2505.13050},
  year={2025}
}

@article{lin2025geocomplete,
  title={Geocomplete: Geometry-aware diffusion for reference-driven image completion},
  author={Lin, Beibei and Chen, Tingting and Tan, Robby T},
  journal={arXiv preprint arXiv:2510.03110},
  year={2025}
}
}

\appendix

\twocolumn[{%
	\maketitle
	\renewcommand\twocolumn[1][]{#1}%
	\begin{center}
		\centering
		\includegraphics[width=\linewidth, height = 0.60\linewidth]{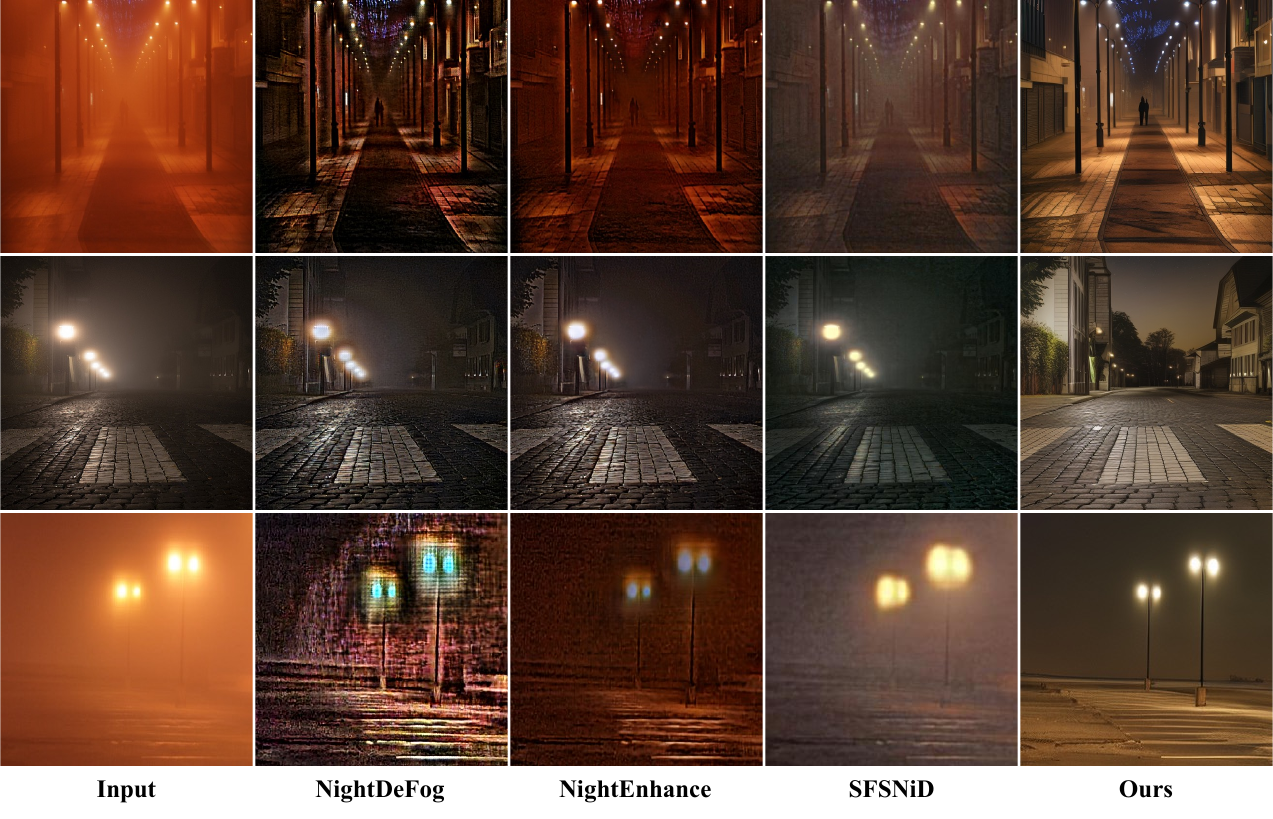}
		\vspace{-7mm}
		\captionof{figure}{Qualitative results from NightDeFog'20~\cite{yan2020nighttime}, NightEnhance'23~\cite{jin2023enhancing}, SFSNiD'24~\cite{cong2024semi} and our method on real-world data. Our method not only reduces dense haze and strong glow but also infers missing background details and content in severely degraded regions.
		}
		
		\label{fig_trailer}
	\end{center}%
}]

\section{More Results on Real-world Datasets}
\begin{figure*}[t]
	\centering
    \subfloat[Input]
	{\includegraphics[width=6cm, height=6cm]{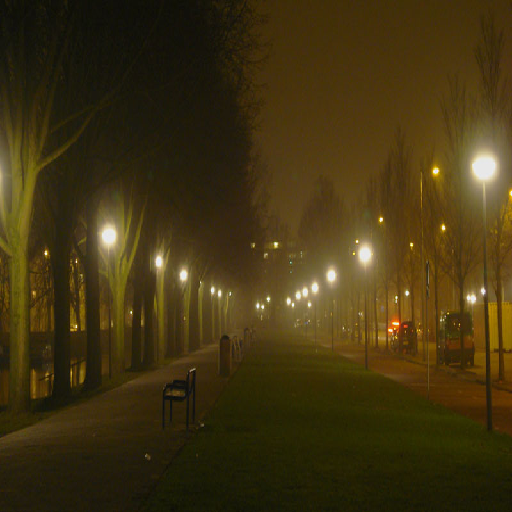}}\hspace{1pt}
    \subfloat[DiT~\cite{Peebles2023DiT}]
	{\includegraphics[width=6cm, height=6cm]{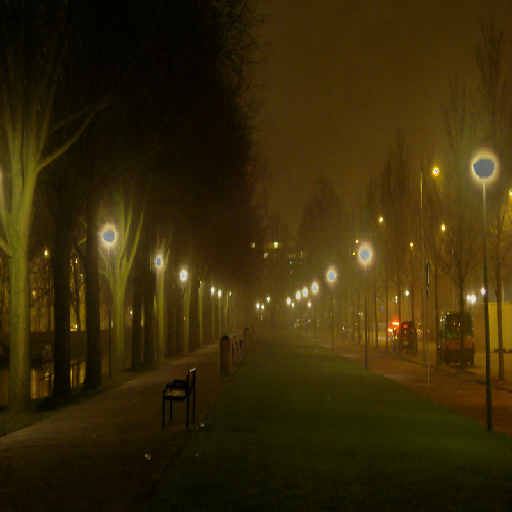}}\hspace{1pt}
    \subfloat[NightDeFog'20~\cite{yan2020nighttime}]
	{\includegraphics[width=6cm, height=6cm]{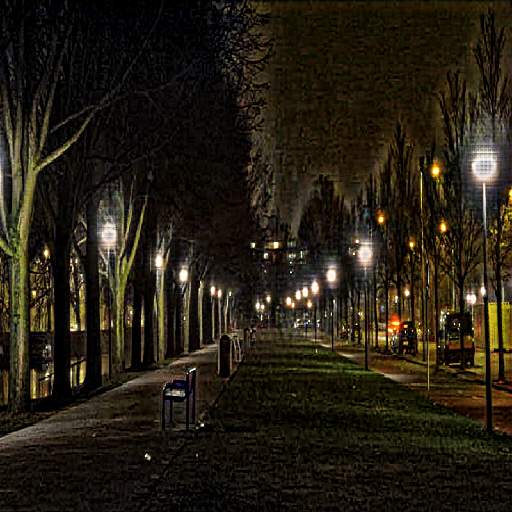}}\hspace{1pt}
     \subfloat[NightEnhance'23~\cite{jin2023enhancing}]
	{\includegraphics[width=6cm, height=6cm]{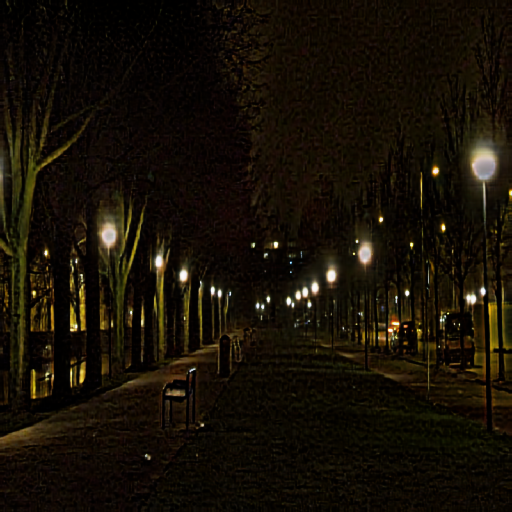}}\hspace{1pt}
    \subfloat[SFSNiD'24~\cite{cong2024semi}]
	{\includegraphics[width=6cm, height=6cm]{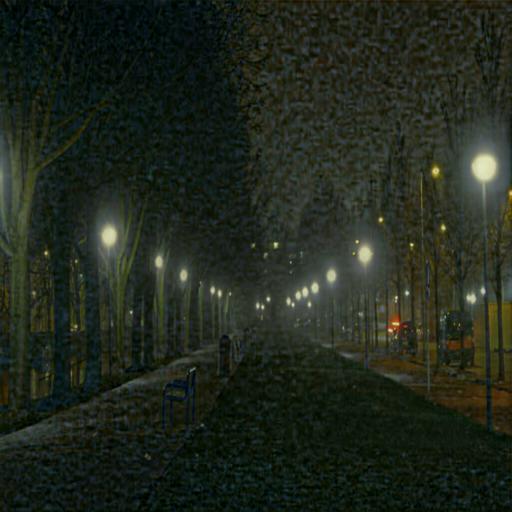}}\hspace{1pt}
    \subfloat[\textbf{Ours}]
	{\includegraphics[width=6cm, height=6cm]{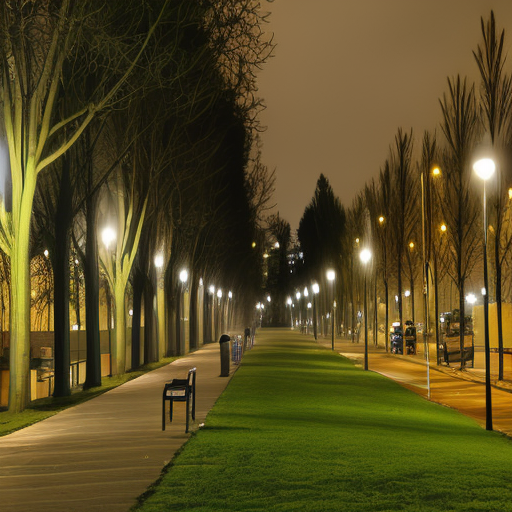}}\hspace{1pt}

	\caption{Qualitative results from DiT'23~\cite{Peebles2023DiT}, 
NightDeFog'20~\cite{yan2020nighttime}, 
NightEnhance'23~\cite{jin2023enhancing}, SFSNiD'24~\cite{cong2024semi}, and our method, on the real-world dataset. Our
method not only reduces dense haze and strong glow effects but also infers missing backgrounds in severely degraded regions. Zoom in
for better visualization.
	}
	\label{fig_exp1}
 \vspace{-6mm}
\end{figure*}

\begin{figure*}[t]
	\centering

    \subfloat[Input]
	{\includegraphics[width=6cm, height=6cm]{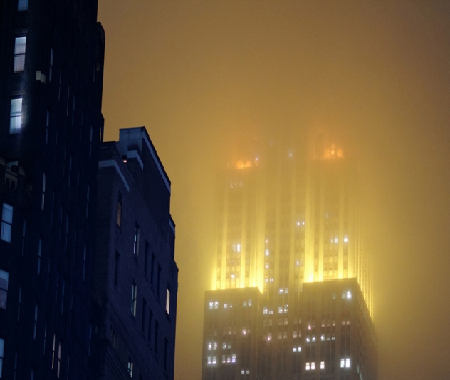}}\hspace{1pt}
    \subfloat[DiT~\cite{Peebles2023DiT}]
	{\includegraphics[width=6cm, height=6cm]{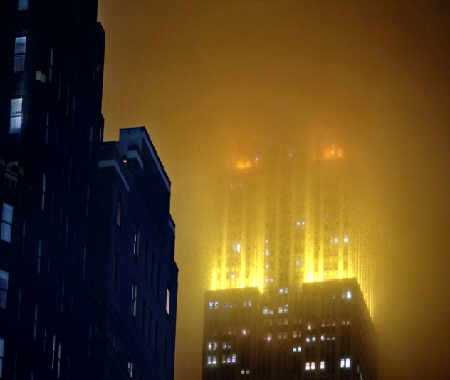}}\hspace{1pt}
    \subfloat[NightDeFog'20~\cite{yan2020nighttime}]
	{\includegraphics[width=6cm, height=6cm]{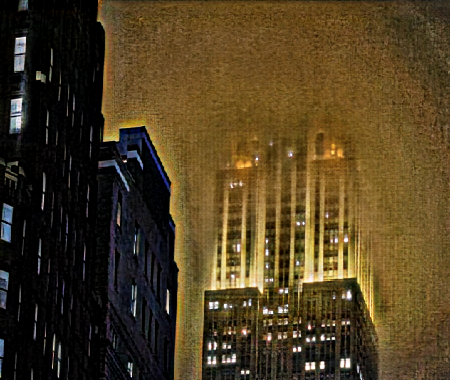}}\hspace{1pt}
     \subfloat[NightEnhance'23~\cite{jin2023enhancing}]
	{\includegraphics[width=6cm, height=6cm]{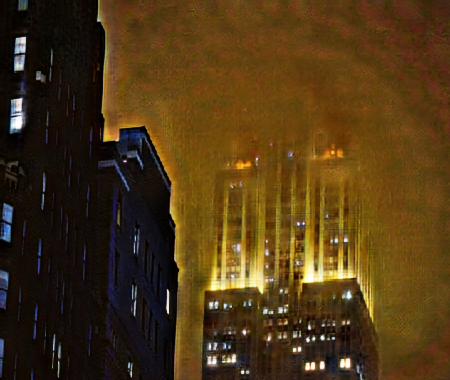}}\hspace{1pt}
    \subfloat[SFSNiD'24~\cite{cong2024semi}]
	{\includegraphics[width=6cm, height=6cm]{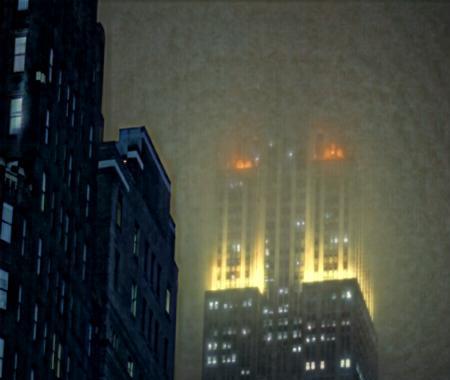}}\hspace{1pt}
    \subfloat[\textbf{Ours}]
	{\includegraphics[width=6cm, height=6cm]{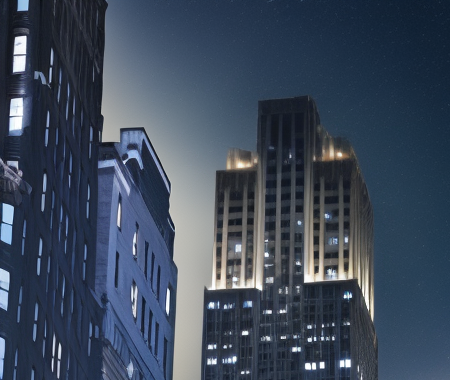}}\hspace{1pt}

	\caption{Qualitative results from DiT'23~\cite{Peebles2023DiT}, 
NightDeFog'20~\cite{yan2020nighttime}, 
NightEnhance'23~\cite{jin2023enhancing}, SFSNiD'24~\cite{cong2024semi}, and our method, on the real-world dataset. Our
method not only reduces dense haze and strong glow effects but also infers missing backgrounds in severely degraded regions. Zoom in
for better visualization.
	}
	\label{fig_exp2}
 \vspace{-6mm}
\end{figure*}

\begin{figure*}[t]
	\centering

    \subfloat[Input]
	{\includegraphics[width=6cm, height=6cm]{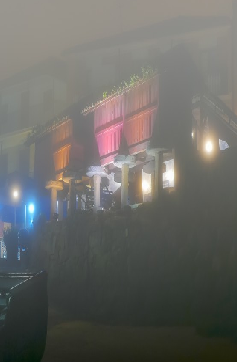}}\hspace{1pt}
    \subfloat[DiT~\cite{Peebles2023DiT}]
	{\includegraphics[width=6cm, height=6cm]{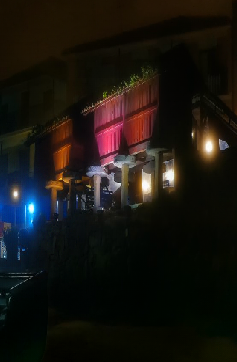}}\hspace{1pt}
    \subfloat[NightDeFog'20~\cite{yan2020nighttime}]
	{\includegraphics[width=6cm, height=6cm]{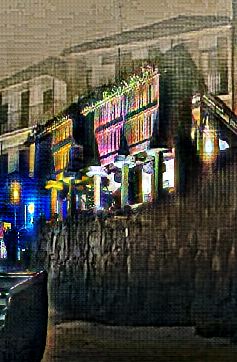}}\hspace{1pt}
     \subfloat[NightEnhance'23~\cite{jin2023enhancing}]
	{\includegraphics[width=6cm, height=6cm]{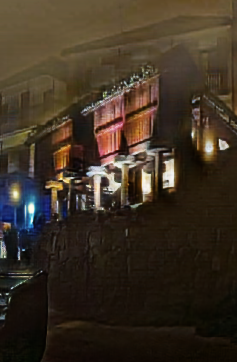}}\hspace{1pt}
    \subfloat[SFSNiD'24~\cite{cong2024semi}]
	{\includegraphics[width=6cm, height=6cm]{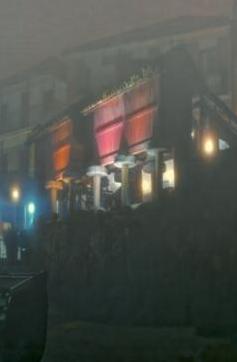}}\hspace{1pt}
    \subfloat[\textbf{Ours}]
	{\includegraphics[width=6cm, height=6cm]{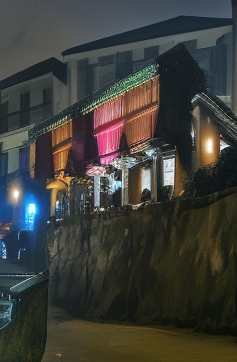}}\hspace{1pt}

	\caption{Qualitative results from DiT'23~\cite{Peebles2023DiT}, 
NightDeFog'20~\cite{yan2020nighttime}, 
NightEnhance'23~\cite{jin2023enhancing}, SFSNiD'24~\cite{cong2024semi}, and our method, on the real-world dataset. Our
method not only reduces dense haze and strong glow effects but also infers missing backgrounds in severely degraded regions. Zoom in
for better visualization.
	}
	\label{fig_exp3}
\end{figure*}

\begin{figure*}[t]
	\centering

    \subfloat[Input]
	{\includegraphics[width=6cm, height=6cm]{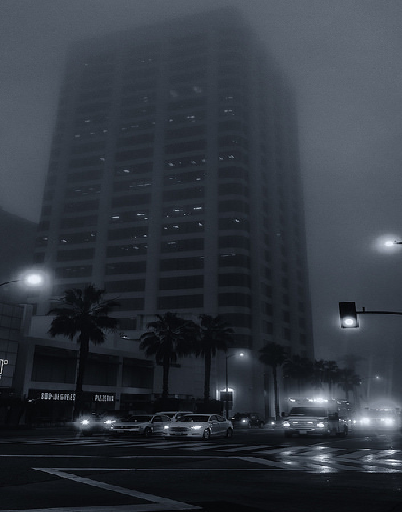}}\hspace{1pt}
    \subfloat[DiT~\cite{Peebles2023DiT}]
	{\includegraphics[width=6cm, height=6cm]{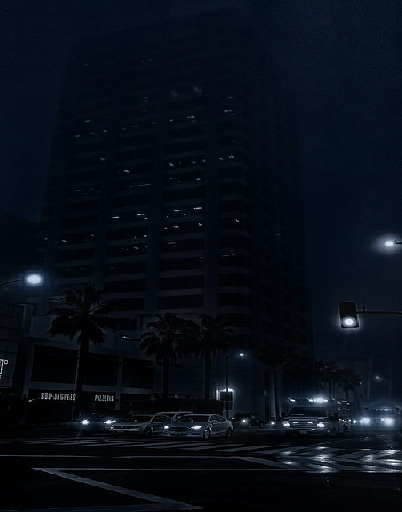}}\hspace{1pt}
    \subfloat[NightDeFog'20~\cite{yan2020nighttime}]
	{\includegraphics[width=6cm, height=6cm]{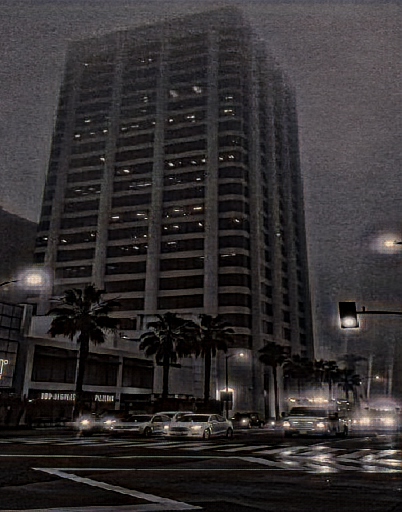}}\hspace{1pt}
     \subfloat[NightEnhance'23~\cite{jin2023enhancing}]
	{\includegraphics[width=6cm, height=6cm]{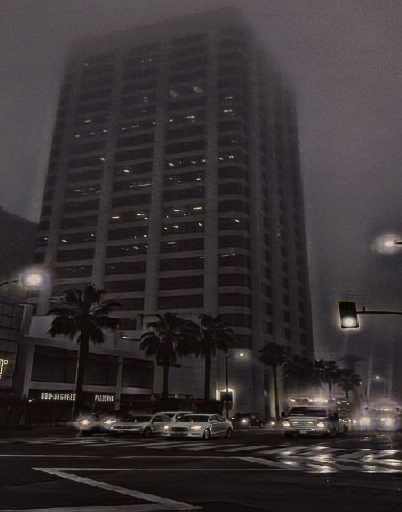}}\hspace{1pt}
    \subfloat[SFSNiD'24~\cite{cong2024semi}]
	{\includegraphics[width=6cm, height=6cm]{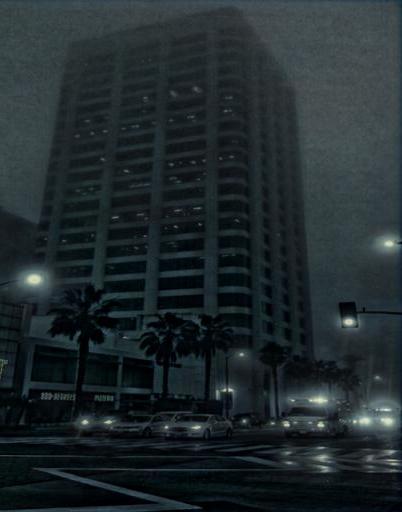}}\hspace{1pt}
    \subfloat[\textbf{Ours}]
	{\includegraphics[width=6cm, height=6cm]{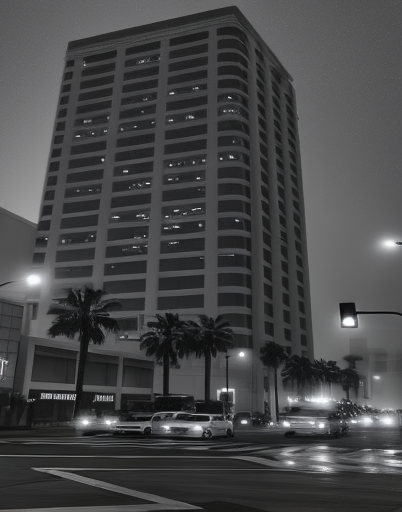}}\hspace{1pt}

	\caption{Qualitative results from DiT'23~\cite{Peebles2023DiT}, 
NightDeFog'20~\cite{yan2020nighttime}, 
NightEnhance'23~\cite{jin2023enhancing}, SFSNiD'24~\cite{cong2024semi}, and our method, on the real-world dataset. Our
method not only reduces dense haze and strong glow effects but also infers missing backgrounds in severely degraded regions. Zoom in
for better visualization.
	}
	\label{fig_exp4}
\end{figure*}

\begin{figure*}[t]
	\centering

    \subfloat[Input]
	{\includegraphics[width=6cm, height=6cm]{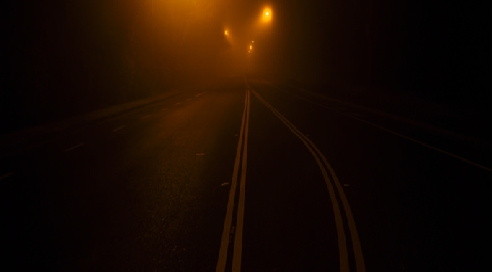}}\hspace{1pt}
    \subfloat[DiT~\cite{Peebles2023DiT}]
	{\includegraphics[width=6cm, height=6cm]{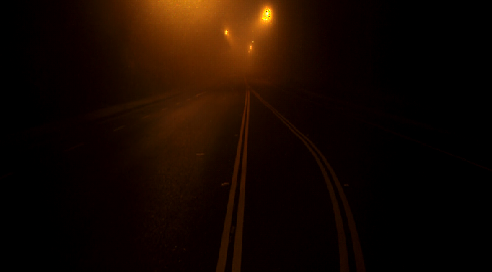}}\hspace{1pt}
    \subfloat[NightDeFog'20~\cite{yan2020nighttime}]
	{\includegraphics[width=6cm, height=6cm]{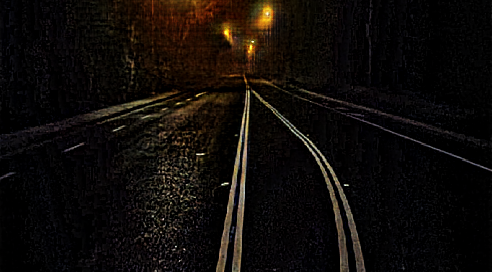}}\hspace{1pt}
     \subfloat[NightEnhance'23~\cite{jin2023enhancing}]
	{\includegraphics[width=6cm, height=6cm]{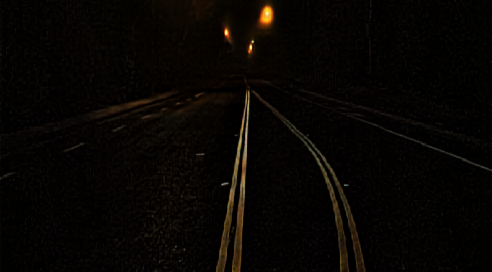}}\hspace{1pt}
    \subfloat[SFSNiD'24~\cite{cong2024semi}]
	{\includegraphics[width=6cm, height=6cm]{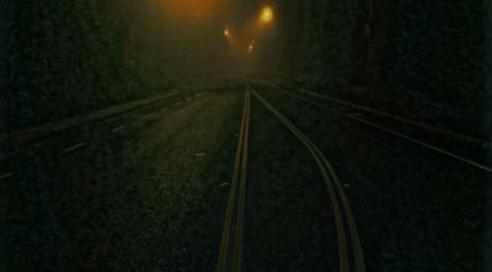}}\hspace{1pt}
    \subfloat[\textbf{Ours}]
	{\includegraphics[width=6cm, height=6cm]{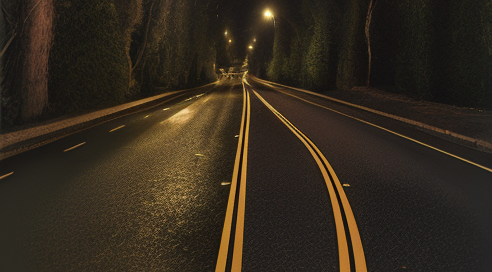}}\hspace{1pt}

	\caption{Qualitative results from DiT'23~\cite{Peebles2023DiT}, 
NightDeFog'20~\cite{yan2020nighttime}, 
NightEnhance'23~\cite{jin2023enhancing}, SFSNiD'24~\cite{cong2024semi}, and our method, on the real-world dataset. Our
method not only reduces dense haze and strong glow effects but also infers missing backgrounds in severely degraded regions. Zoom in
for better visualization.
	}
	\label{fig_exp5}
\end{figure*}

\begin{figure*}[t]
	\centering

    \subfloat[Input]
	{\includegraphics[width=6cm, height=6cm]{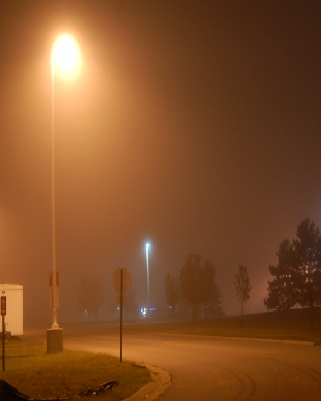}}\hspace{1pt}
    \subfloat[DiT~\cite{Peebles2023DiT}]
	{\includegraphics[width=6cm, height=6cm]{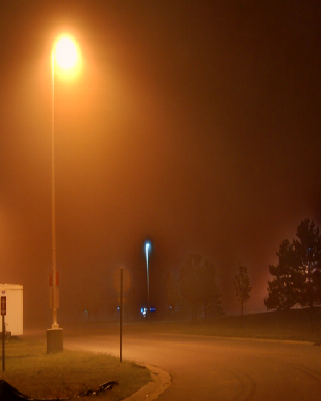}}\hspace{1pt}
    \subfloat[NightDeFog'20~\cite{yan2020nighttime}]
	{\includegraphics[width=6cm, height=6cm]{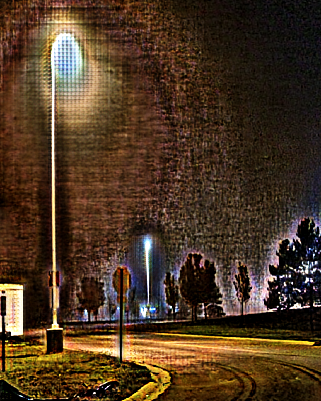}}\hspace{1pt}
     \subfloat[NightEnhance'23~\cite{jin2023enhancing}]
	{\includegraphics[width=6cm, height=6cm]{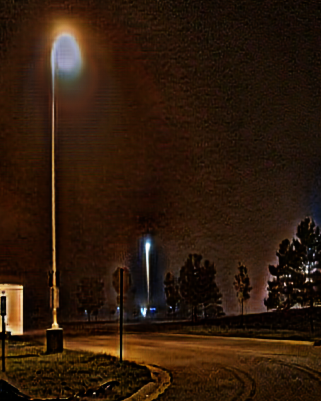}}\hspace{1pt}
    \subfloat[SFSNiD'24~\cite{cong2024semi}]
	{\includegraphics[width=6cm, height=6cm]{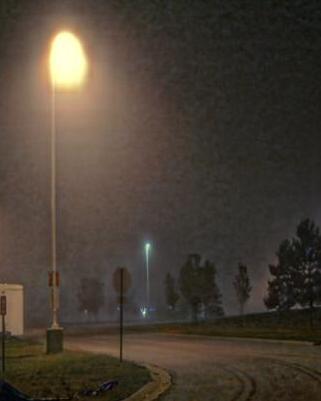}}\hspace{1pt}
    \subfloat[\textbf{Ours}]
	{\includegraphics[width=6cm, height=6cm]{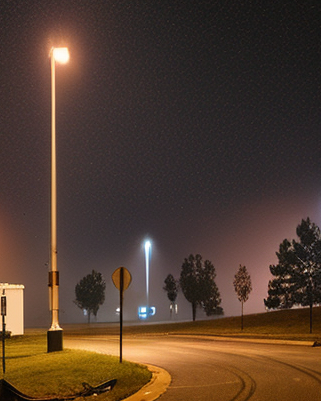}}\hspace{1pt}

	\caption{Qualitative results from DiT'23~\cite{Peebles2023DiT}, 
NightDeFog'20~\cite{yan2020nighttime}, 
NightEnhance'23~\cite{jin2023enhancing}, SFSNiD'24~\cite{cong2024semi}, and our method, on the real-world dataset. Our
method not only reduces dense haze and strong glow effects but also infers missing backgrounds in severely degraded regions. Zoom in
for better visualization.
	}
	\label{fig_exp6}
\end{figure*}

\begin{figure*}[t]
	\centering

    \subfloat[Input]
	{\includegraphics[width=6cm, height=6cm]{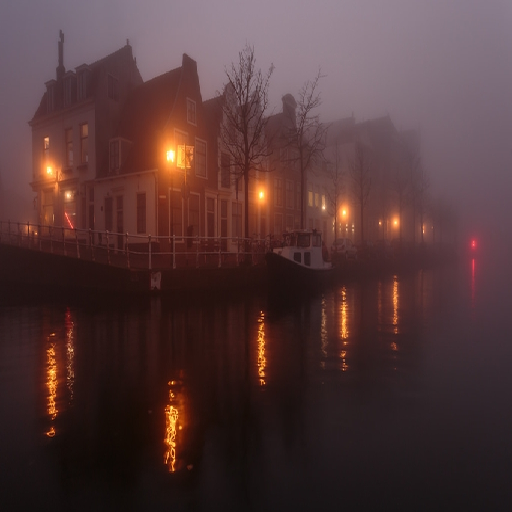}}\hspace{1pt}
    \subfloat[DiT~\cite{Peebles2023DiT}]
	{\includegraphics[width=6cm, height=6cm]{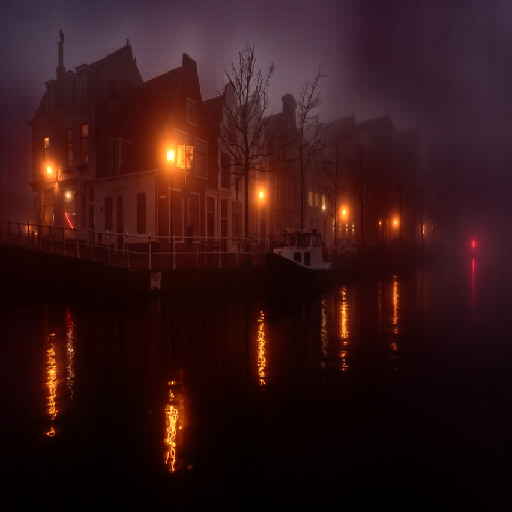}}\hspace{1pt}
    \subfloat[NightDeFog'20~\cite{yan2020nighttime}]
	{\includegraphics[width=6cm, height=6cm]{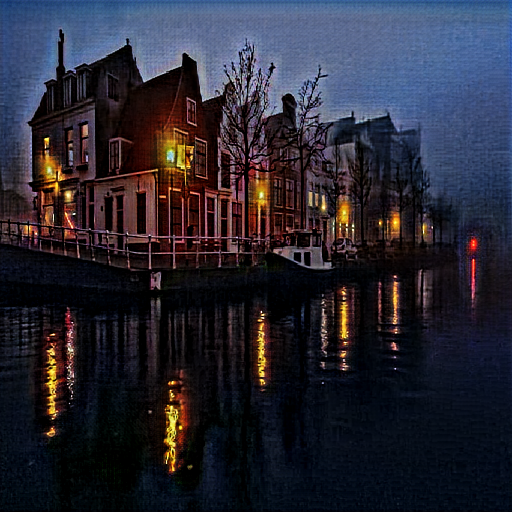}}\hspace{1pt}
     \subfloat[NightEnhance'23~\cite{jin2023enhancing}]
	{\includegraphics[width=6cm, height=6cm]{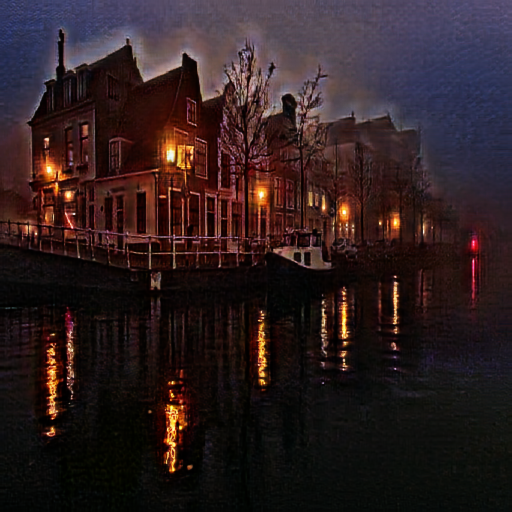}}\hspace{1pt}
    \subfloat[SFSNiD'24~\cite{cong2024semi}]
	{\includegraphics[width=6cm, height=6cm]{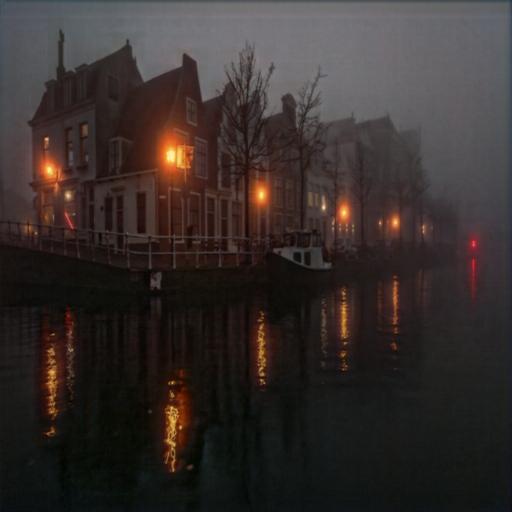}}\hspace{1pt}
    \subfloat[\textbf{Ours}]
	{\includegraphics[width=6cm, height=6cm]{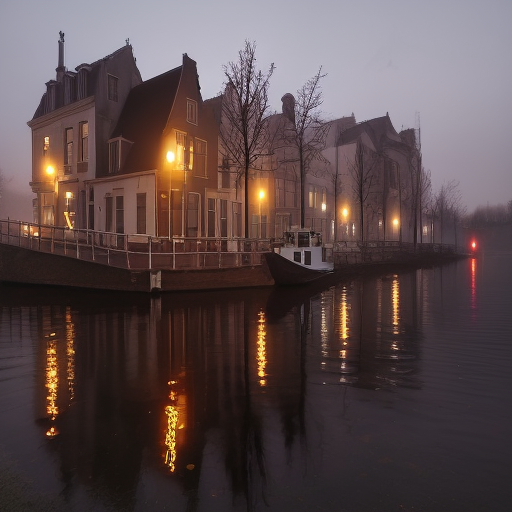}}\hspace{1pt}

	\caption{Qualitative results from DiT'23~\cite{Peebles2023DiT}, 
NightDeFog'20~\cite{yan2020nighttime}, 
NightEnhance'23~\cite{jin2023enhancing}, SFSNiD'24~\cite{cong2024semi}, and our method, on the real-world dataset. Our
method not only reduces dense haze and strong glow effects but also infers missing backgrounds in severely degraded regions. Zoom in
for better visualization.
	}
	\label{fig_exp7}
 \vspace{-3mm}
\end{figure*}

In this section, we present additional real-world dehazing results comparing our \textit{BeyondHaze} with state-of-the-art (SOTA) methods, including DiT'23~\cite{Peebles2023DiT}, NightDeFog'20~\cite{yan2020nighttime}, NightEnhance'23~\cite{jin2023enhancing}, and SFSNiD'24~\cite{cong2024semi}. DiT'23 serves as a general image restoration backbone trained on synthetic datasets, while the other SOTA methods are specialized nighttime image dehazing approaches trained using both labeled synthetic datasets and unlabeled real-world datasets.
The experimental results are shown in Figs.~\ref{fig_trailer}, \ref{fig_exp1}, 
\ref{fig_exp2}, \ref{fig_exp3}, \ref{fig_exp4}, \ref{fig_exp5}, 
\ref{fig_exp6}, and \ref{fig_exp7}. 
We can find that our \textit{BeyondHaze} not only reduces dense haze and strong glow effects but also infers missing backgrounds in severely degraded regions.

\begin{figure*}[t!]
	\centering
    \includegraphics[width=1\linewidth]{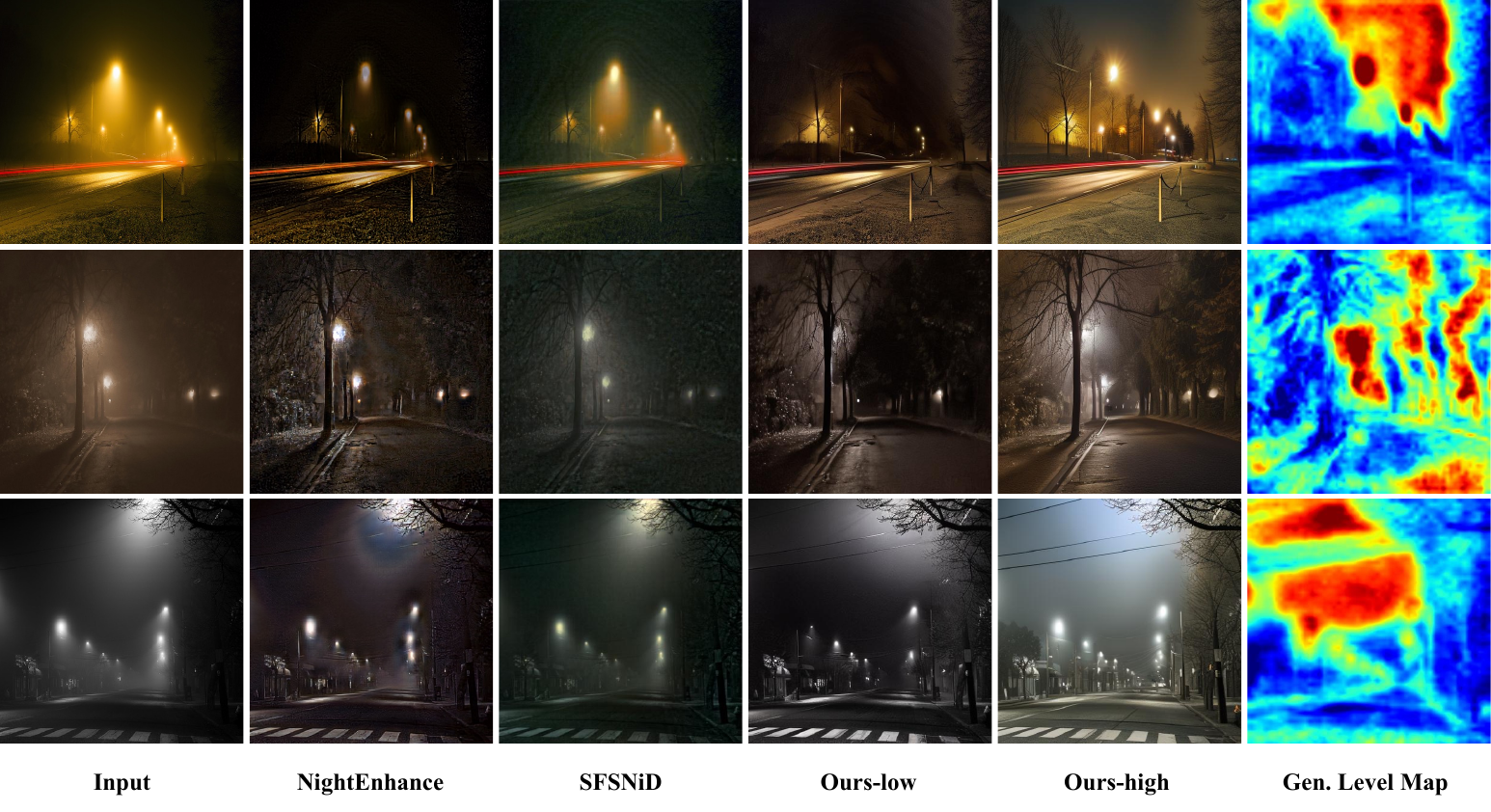}
    \vspace{-8mm}
	\caption{Qualitative results from NightEnhance’23~\cite{jin2023enhancing}, SFSNiD’24~\cite{cong2024semi}, and our method on real-world datasets. “Ours-low” refers to our low-generative results, which prioritize haze removal, while “Ours-high” refers to our high-generative results, allowing the network to infer details and backgrounds in severely degraded regions. “Gen. Level Map” shows the generative-level maps of “Ours-high,” with red indicating high and blue indicating low generative regions. Zoom in for better visualization.
	}
	\label{figs_exp1_gen_map}
\end{figure*}

\section{More Results with Generative Level Maps}
In this section, we provide more visualizations of our generative level maps in Figure~\ref{figs_exp1_gen_map}. The red-highlighted areas in the map indicate regions likely to contain hallucinated content, such as the details of trees and the color and brightness of sky regions.

\section{Experimental Details} 
Our BeyondHaze framework is built upon generative dehazing priors and controllable generative dehazing. We provide additional training details as follows. 

\subsection{Generative Dehazing Priors} 

\paragraph{Pre-trained Dehazing Model Details}
Our pre-trained dehazing model is a ViT-Large \cite{he2022masked} architecture with 24 encoder layers and 8 decoder layers. During training, we initialize the model with pre-trained parameters from \cite{he2022masked} and fine-tune it using strong augmentations from \cite{lin2024nighthaze}, where clear images are intentionally degraded with light effects and noise. This approach forces the model to learn robust background priors, enabling it to infer clear scene information despite degraded inputs. 

During training, input images are randomly cropped to 224 × 224. We set the total number of training steps to 20,000 and use a batch size of 128. The Adam optimizer is used with an initial learning rate of 1.5e-4. During inference, we apply overlapping sampling to reconstruct high-resolution images, as ViT-Large only supports an input size of 224 × 224. Specifically, a 224 × 224 sliding window is used to extract overlapping regions from high-resolution hazy inputs. The pre-trained dehazing model is then applied to restore these regions. Since each pixel receives multiple predictions, we average them to obtain the final dehazing results. 

\paragraph{Details of Initial Dehazing Pairs} Once the pre-trained dehazing model is obtained, we utilize it to construct training pairs from real-world data. These pairs, along with their text prompts, are then used to fine-tune the diffusion model.

\begin{figure*}[t]
	\centering
	{\includegraphics[width=4.5cm, height=1.5cm]{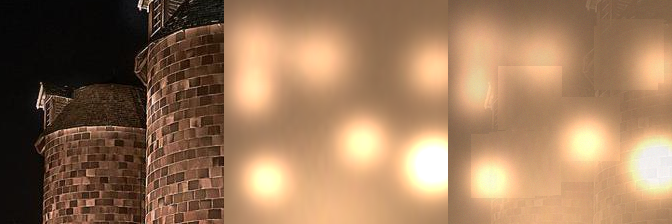}}\hspace{1pt}
	{\includegraphics[width=4.5cm, height=1.5cm]{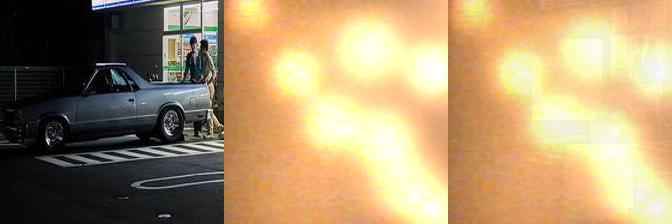}}\hspace{1pt}
	{\includegraphics[width=4.5cm, height=1.5cm]{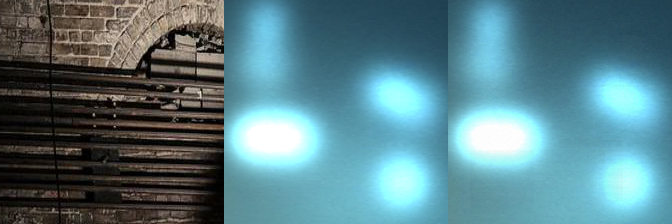}}\hspace{1pt}
    \\
    
	{\includegraphics[width=4.5cm, height=1.5cm]{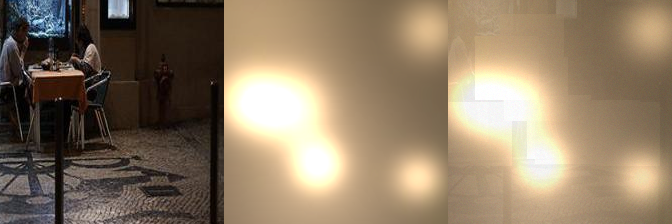}}\hspace{1pt}
	{\includegraphics[width=4.5cm, height=1.5cm]{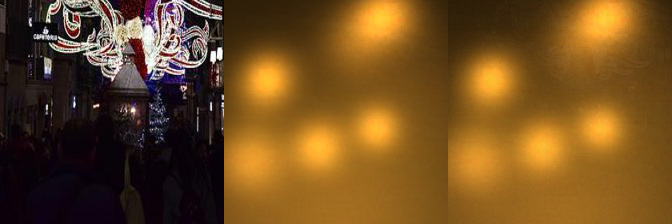}}\hspace{1pt}
	{\includegraphics[width=4.5cm, height=1.5cm]{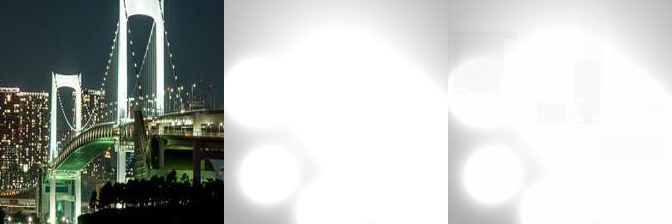}}\hspace{1pt}
    \\

	{\includegraphics[width=4.5cm, height=1.5cm]{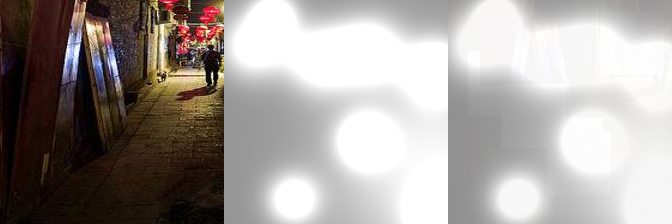}}\hspace{1pt}
	{\includegraphics[width=4.5cm, height=1.5cm]{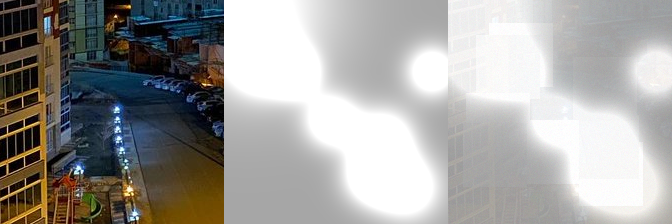}}\hspace{1pt}
	{\includegraphics[width=4.5cm, height=1.5cm]{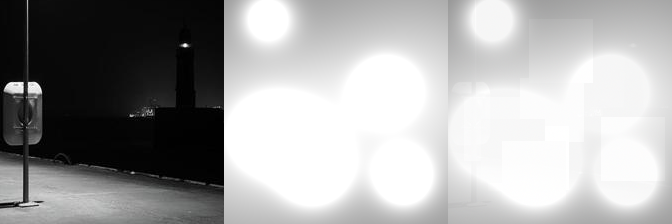}}\hspace{1pt}
    \\
    
	{\includegraphics[width=4.5cm, height=1.5cm]{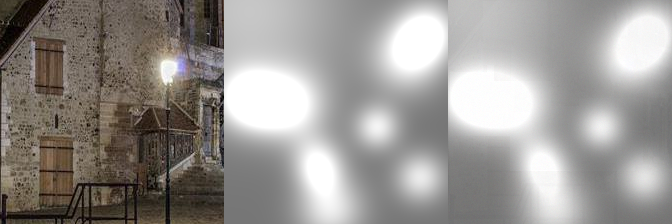}}\hspace{1pt}
	{\includegraphics[width=4.5cm, height=1.5cm]{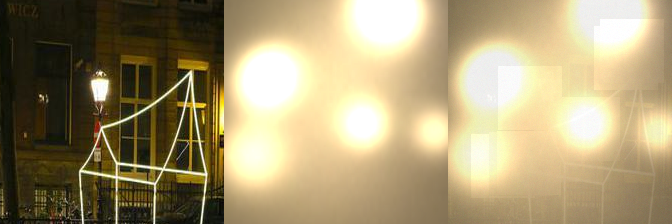}}\hspace{1pt}
	{\includegraphics[width=4.5cm, height=1.5cm]{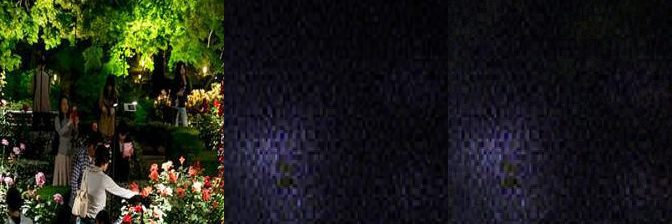}}\hspace{1pt}
    \\

\caption{Examples of our severely augmented training pairs. For each triplet, we show the ground-truth image (left), the augmented light map (middle), and the augmented image (right).}

	\label{fig_training_vis}
\end{figure*}

\subsection{Controllable Generative Dehazing} 

\subsubsection{Detail Enhancement Model} During training, we use the detail enhancement model \cite{flux1} to recover fine-scale details in the initial dehazed images produced by the pre-trained dehazing model, resulting in detail-enhanced image pairs. This model is built on FLUX.1 \cite{flux1} and incorporates a ControlNet module trained using the augmentation strategy from \cite{wang2021real}.

\subsubsection{Severe Degradation Model} Our severe degradation model heavily degrades clear nighttime images by blending light effects and injecting noise, producing a strongly augmented dataset formulated as in \cite{lin2024nighthaze}: 
\begin{equation}
    \label{eqn_generate_model}
    I = W_{b}*J + (1-W_{b})*L + \epsilon,
\end{equation}
where $I$ is the augmented image, $J$ is the clear image, $W_{b}$ is the blend weight map, $L$ is the light map and  $\epsilon$ is the noise. 
We define the severity of the augmentation \( S \) as: 
\begin{equation}
    S = 1 - \mathbb{E} [\frac{ W_b \cdot J }{ I }]
\end{equation}
where \( \mathbb{E}[\cdot] \) represents the mean over all spatial dimensions (height, width, and channels) and $S$ defines the severity ratio.

\vspace{-2mm}
\paragraph{Blend Weight Map $W_{b}$} During training, for each input clear image \( J \), we randomly sample a value from a uniform distribution \( (0.001, 0.1) \) to initialize a \( 512 \times 512 \) map. Next, we randomly select eight \( 128 \times 128 \) regions from this map. For each selected region, we randomly adjust its values within the range \( (0, 0.04) \). As a result, the blend weight map becomes non-uniform. 

\vspace{-2mm}
\paragraph{Light Map $L$} We use real-world light maps for the augmentation, which is sourced from \cite{lin2024nighthaze}. At each training step, we randomly select a light map for each clear image.  Then, we select various regions and amplify their brightness using Gaussian kernels. The number of selected regions is between 1 and 10, and the size of Gaussian kernels ranges from 15 to 160. 

\vspace{-2mm}
\paragraph{Noise $\epsilon$} We introduce Gaussian noise to augment input images. We set \( W_n \) to 0.1, resulting in a noise range of \( (0, 0.03) \).  

With these parameters, the severity \( S \) averages approximately 98\% over all training steps.  
Note that the purpose of our severe augmentation model is to heavily degrade clear nighttime images, simulating real-world scenarios where background details are significantly degraded or entirely lost. It is not intended to perfectly simulate glow/haze effects or transmission maps. Therefore, Gaussian noise, the blend weight map, and the light map are sufficient as long as the augmentation remains severe. Figure~\ref{fig_training_vis} visualizes examples of the severely augmented training pairs.

\subsubsection{Training Details}
The customized text prompt for initial dehazing pairs \(t_{\rm {ID}}\) is ``\textit{A dehazed image with slight degradation.}", while the text prompt for detail enhancement pairs \(t_{\rm {DE}}\) is ``\textit{A high-resolution, dehazed image with slight degradation.}". The prompt for background synthesis pairs \(t_{\rm {bs}}\) is ``\textit{An image with no degradation, generation.}". Since the only difference between the text prompts for initial-dehazing and detail enhancement pairs is the term "high-resolution," this distinction can be effectively learned by our diffusion model. 

In each training batch, we sample paired data equally from the initial-dehazing, detail-enhancement, and background-synthesis subsets to ensure balanced learning. 
The total training batch size is set to 12, with a gradient accumulation step of 4 for stable training. The learning rate is set to \( 2 \times 10^{-4} \) for the U-Net, with an input resolution of \( 512 \times 512 \). The total number of training steps is 10,000. All experiments are conducted using 4 NVIDIA GPUs, each with 24 GB of memory. 

During inference, given a real-world nighttime haze image, we utilize the fine-tuned diffusion model to enhance its visibility. With the high-generative prompt \(t_{\text{high}}\), \textit{``A high-resolution, dehazed image with no degradation, generation''}, the model produces highly generative dehazed results with synthesized details and background content. Conversely, the low-generative prompt \(t_{\text{low}}\), \textit{``A dehazed image with slight degradation''}, suppresses generative synthesis, yielding results with higher fidelity but relatively lower visual quality.

\section{Ablation Studies}
\begin{figure*}[t!]
	\centering
	\includegraphics[width=1\linewidth]{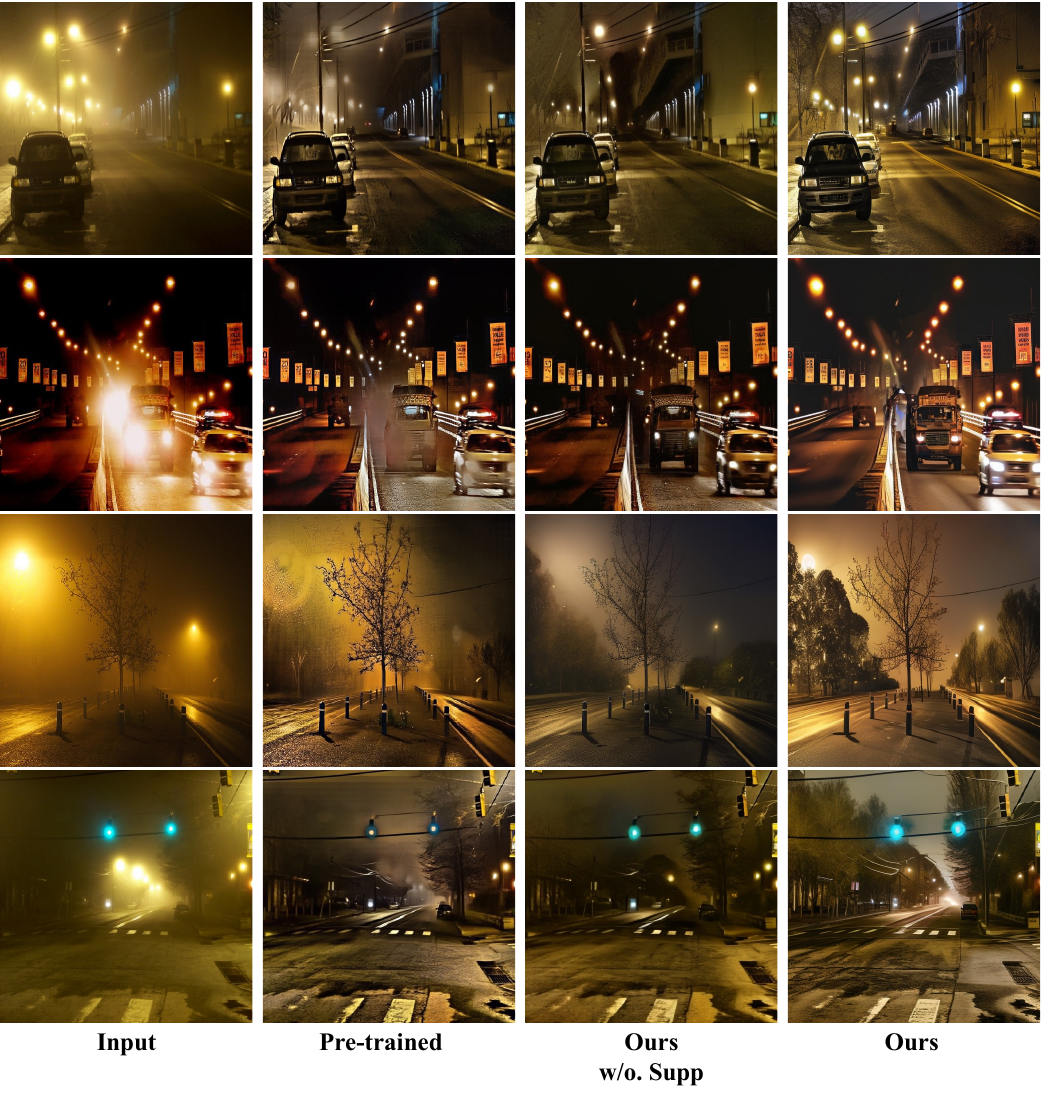}
    \vspace{-6mm}
	\caption{
    Ablation study on real-world haze datasets.
    “Pre-trained” denotes the output of our pre-trained dehazing model. “Ours w/o supp” uses only our generative dehazing priors. “Ours” combines both our generative priors and supplementary components.
    }
	\label{figs_abl_vis01}
\end{figure*}

In this section, we present visual ablation results to validate the effectiveness of our generative dehazing priors and the supplementary components. The comparisons are shown in Figure~\ref{figs_abl_vis01}. 
While our pre-trained model learns strong background priors, its outputs still exhibit noticeable artifacts, especially in regions with dense haze or intense glow. In contrast, \textit{Ours w/o supp} equips the diffusion model with both dehazing and generative capabilities, resulting in more realistic outputs with enhanced visibility and improved background recovery. Finally, \textit{Ours}, which incorporates both generative priors and supplementary training pairs, further enhances fine-scale details and restores severely obscured scene content. These results highlight the benefits of guided diffusion training and demonstrate the effectiveness of our user-controllable generative dehazing framework.

\section{Limitations}

Although our method develops generative dehazing capabilities to handle dense haze and glow scenarios, its performance is inherently constrained by the pretrained diffusion model (e.g., Stable Diffusion). For instance, if a scene contains fine-grained elements such as text, signage, or uncommon objects that are underrepresented in the model's training data, the generated content in heavily obscured regions may be less accurate or semantically inconsistent.

\section{Societal Impact}

Nighttime visibility is often degraded by haze and glow effects, which limits the performance of applications such as autonomous driving and surveillance. Our method addresses these challenges by leveraging generative dehazing priors to restore or infer obscured content. This enhancement of nighttime visual quality can benefit downstream vision tasks and improve human interpretation in low-light, haze-affected environments. 

\end{document}